%% file: main.tex
\RequirePackage[svgnames]{xcolor}
 \RequirePackage[svgnames]{xcolor}

\documentclass[11pt,letterpaper]{mystyle}

\usepackage[all]{hypcap}
\usepackage[svgnames]{xcolor}
\usepackage[table]{xcolor}
\usepackage[comma,authoryear,compress]{natbib}
\usepackage{hyperref}[citecolor=lightblue]

\hypersetup{
    colorlinks = true,
    citecolor = {SEUGreenDark},
    linkcolor = {SEUGreenDark},
    urlcolor = {SEUGreen},
}

\usepackage{algorithm}
\usepackage{algorithmicx}
\usepackage{algpseudocode}
\usepackage{microtype}
\usepackage{graphicx}
\expandafter\def\csname ver@subfig.sty\endcsname{}
\usepackage{booktabs} %
\usepackage{tabularx} %
\usepackage{array} %
\usepackage{float}
\usepackage{bigstrut}

\usepackage{amsmath}
\usepackage{amssymb}
\usepackage{mathtools}
\usepackage{amsthm}
\usepackage{mathrsfs}
\usepackage{nicefrac}
\usepackage{dsfont}
\usepackage{enumitem}
\usepackage{subcaption}
\usepackage{graphicx,subfig}
\usepackage{cleveref}

\usepackage{float}

\usepackage[utf8]{inputenc} %
\usepackage[T1]{fontenc}    %
\usepackage{hyperref}       %
\usepackage{url}            %
\usepackage{booktabs}       %
\usepackage{amsfonts}       %
\usepackage{nicefrac}       %
\usepackage{microtype}      %
\usepackage{graphicx}
\usepackage{subcaption} 
\usepackage{amssymb}
\usepackage{fdsymbol}
\usepackage{wrapfig}
\usepackage{lipsum}
\usepackage{enumitem}
\usepackage{stackengine}
\usepackage[font=small,labelfont=bf]{caption}
\usepackage{color}
\usepackage{adjustbox}

\usepackage{rotating}
\usepackage{makecell}

\input{macro}

\newtcolorbox{AIbox}[2][]{aibox,title=#2,#1}
\definecolor{lightblue}{rgb}{0.22,0.45,0.70}%
\definecolor{Gray}{gray}{0.95}
\definecolor{Cornsilk}{rgb}{1.0, 0.97, 0.86}

\definecolor{SafeGreen}{HTML}{BDDE93}
\definecolor{SafePink}{HTML}{FF82A2}
\definecolor{SafeBlue}{HTML}{63BAD9}
\definecolor{SafeOrange}{HTML}{FE8F29}
\definecolor{SafeGold}{HTML}{FAE593}

\newcommand{\boxedcell}[2]{%
  \begingroup
  \setlength{\fboxsep}{1.2pt}%
  \setlength{\fboxrule}{0.35pt}%
  \fcolorbox{white}{#1}{\strut\textbf{#2}}%
  \endgroup
}

\newcommand{\best}[1]{\boxedcell{SafeBlue!20}{#1}}
\newcommand{\secondbest}[1]{\boxedcell{SafePink!20}{#1}}

\usepackage{amsmath}

\usepackage[all]{hypcap}

\title{\textbf{AI-generated Images Challenge Visual Trust in High-risk Scenarios}}

\runningtitle{\textit{Seeing Is No Longer Believing:} AI-generated Images Challenge Visual Trust in High-risk Scenarios}

\author{Yi-Zhi Wang$^{1,2}$, Yichen Xiao$^{1,2}$, Linan Yue$^{1,2}$, Weibo Gao$^{3}$, Yichao Du$^{4}$, Pengfei Fang$^{1,2}$,\\ Shimin Di$^{1,2}$, and Min-Ling Zhang$^{1,2}$}

\affil[1]{School of Computer Science and Engineering, Southeast University}
\affil[2]{Key Laboratory of Computer Network and Information Integration (Southeast University), Ministry of Education}
\affil[3]{The Hong Kong Polytechnic University}
\affil[4]{School of Artificial Intelligence, Wuhan University}

\correspondingauthor{Linan Yue, Email: \href{lnyue@seu.edu.cn}{lnyue@seu.edu.cn}}

\begin{document}

\input{sections/abstract}
\maketitle

\begin{figure}[h]
    \centering
    \begin{subfigure}{0.3\linewidth}
        \centering
        \includegraphics[width=\linewidth]{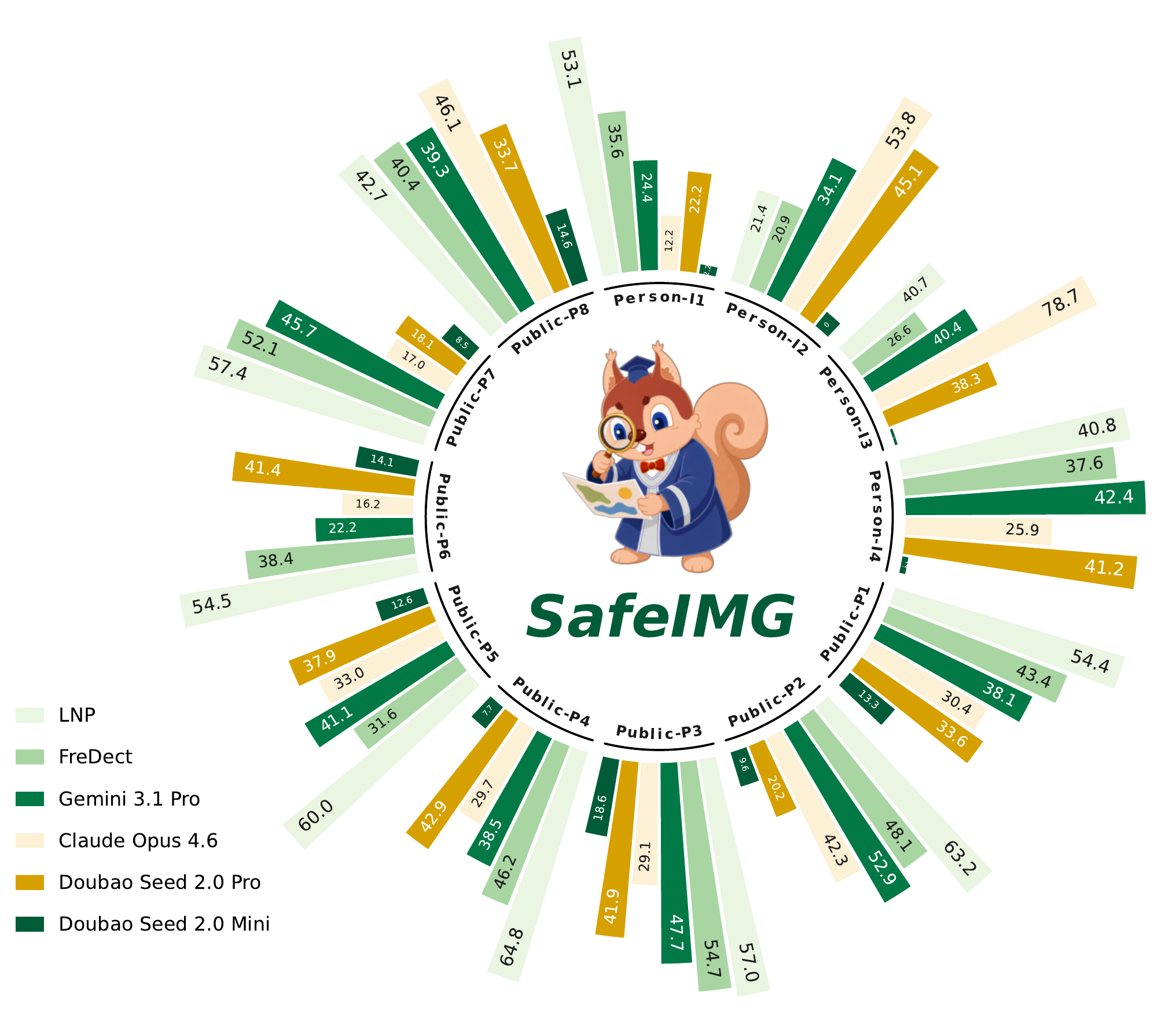}
        \caption{Category-level image detection performance on SafeIMG.}
        \label{fig:radient}
    \end{subfigure}
    \hspace{0.6cm}
    \begin{subfigure}{0.25\linewidth}
        \centering
        \includegraphics[width=\linewidth]{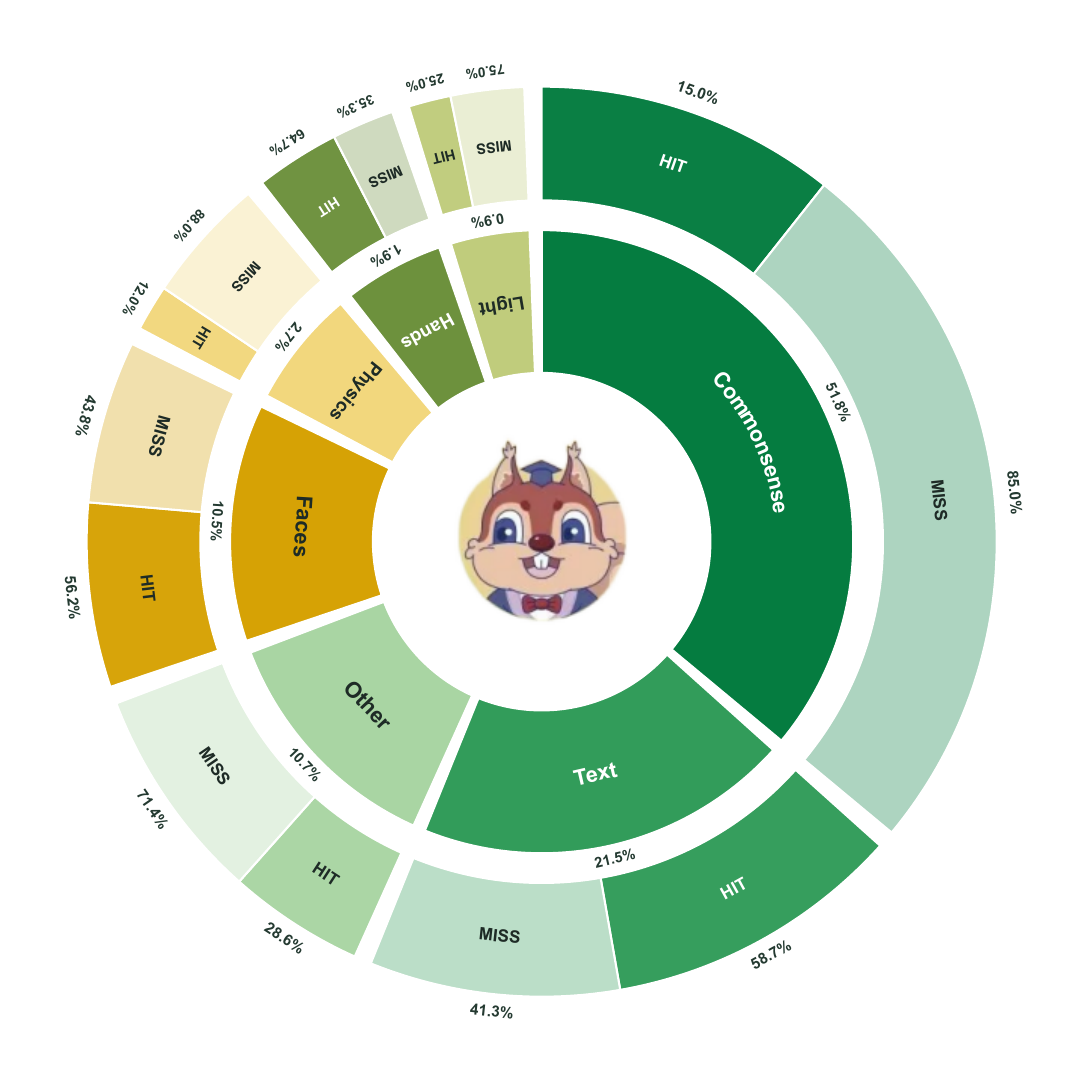}
        \caption{Distribution of human annotated artifact types and hit rates.}
        \label{fig:doublepie}
    \end{subfigure}
    \hspace{0.6cm}
    \begin{subfigure}{0.25\linewidth}
        \centering
        \includegraphics[width=4.1cm]{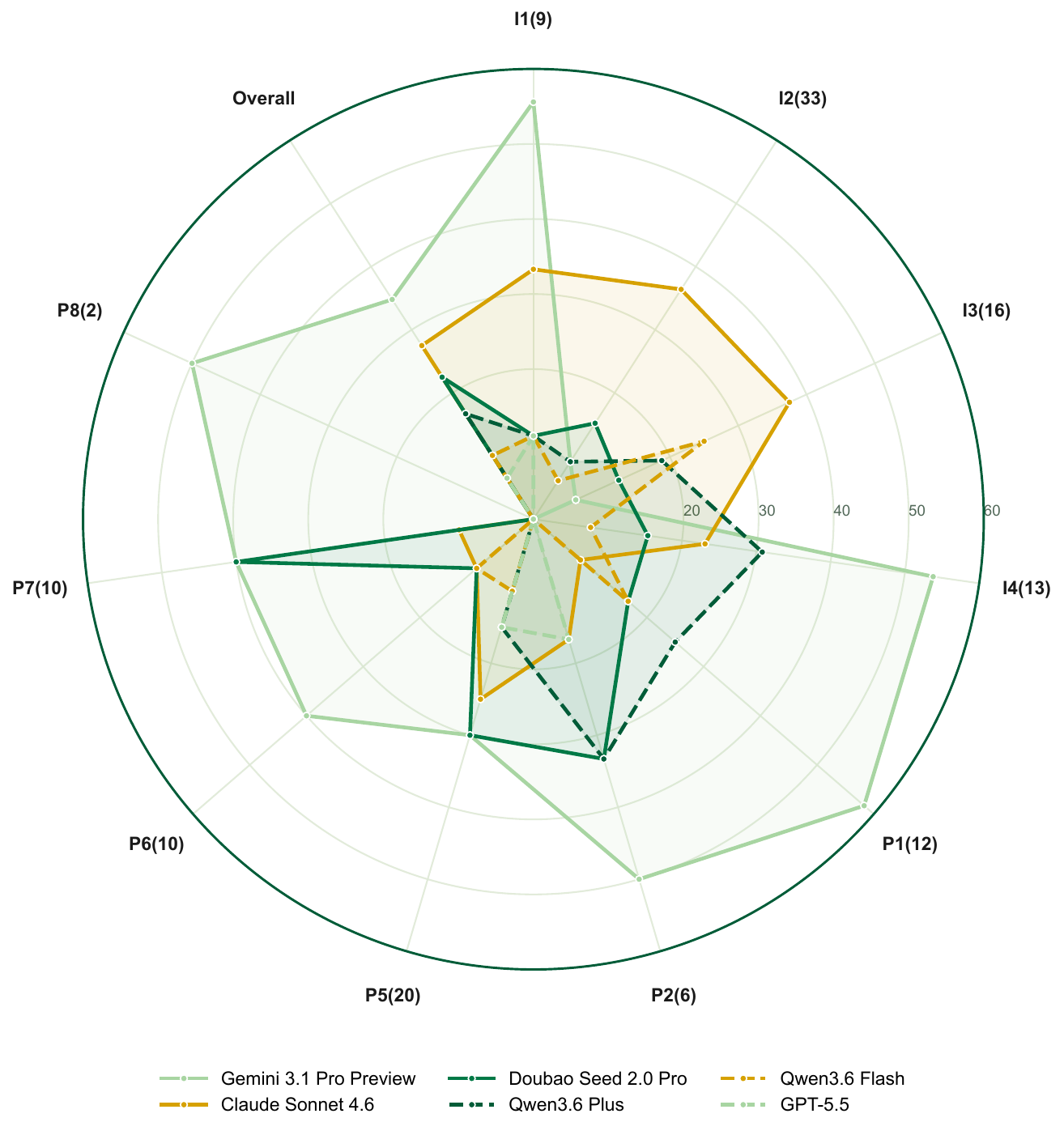}
        \caption{Comparison of representative models on difficult subset.}
        \label{fig:radar}
    \end{subfigure}
    \caption{Overview of SafeIMG and key empirical findings. The figure highlights the challenges of detecting AI-generated images in high-risk scenarios, where current models exhibit uneven capability and limited understanding of human-perceived artifacts.}
    \label{fig:model-detection}
\end{figure}
\section{Introduction}

Image generation models have advanced rapidly in recent years~\citep{ho2020denoising,song2021scorebased,lipman2023flow,esser2024scaling,chen2024pixartsigma,yan2025gptimgeval,chen2025gpt4oimage}. Modern text-to-image systems can synthesize realistic images, follow detailed natural-language instructions, and compose complex visual scenes with fine-grained control over objects, layout, style, and text~\citep{zhang2023adding,tuo2023anytext,esser2024scaling}. These capabilities are no longer limited to professional design software or closed research systems. Public products and API services now allow non-specialist users to generate large volumes of realistic images at low cost by writing natural-language prompts.

This progress changes the security implications of synthetic images. Earlier AI-generated images were often discussed as aesthetic or creative artifacts, where realism and visual quality were the main concerns~\citep{heusel2017gans,karras2019style,dhariwal2021diffusion}. Recent models can increasingly produce images with evidentiary attributes. These images appear to document real-world facts, events, identities, transactions, or records~\citep{chesney2019deep,vaccari2020deepfakes,nightingale2022ai}. For example, a generated image may resemble a news photograph from a disaster scene, a bank-transfer screenshot, an identity credential, a legal document, or a private communication record. In such cases, the image is not merely consumed as visual content. It may shape public judgment, financial trust, identity verification, legal interpretation, or personal reputation. As synthetic images enter these high-risk contexts, the social assumption that ``\textit{seeing is believing}'' becomes increasingly fragile~\citep{chesney2019deep,vaccari2020deepfakes,nightingale2022ai}.

\begin{figure}[t]
    \centering·
    \includegraphics[width=17.5cm]{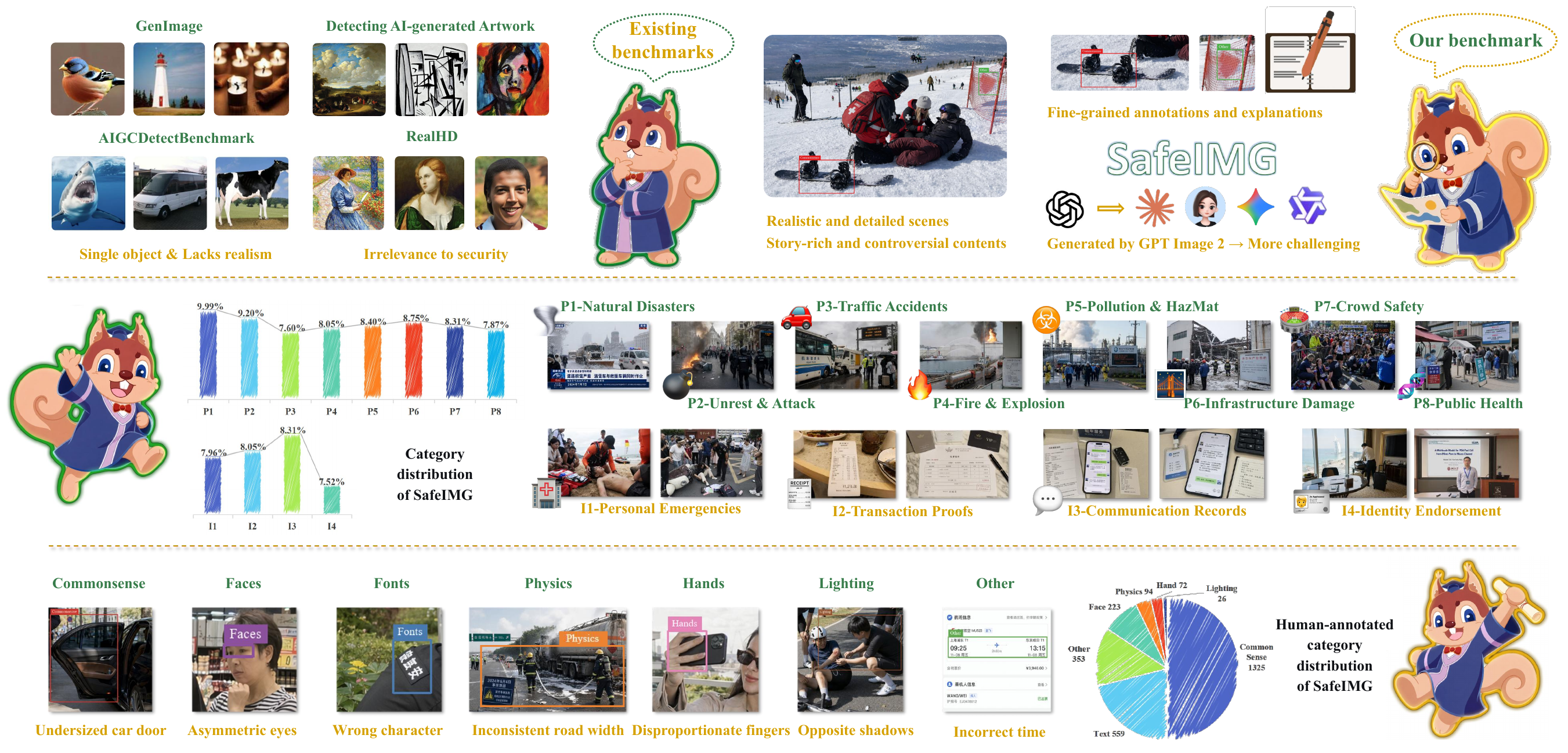}
\caption{Overview and distinguishing characteristics of SafeIMG. SafeIMG organises 12 safety-oriented scenarios into public- and individual-safety domains and uses structured prompts to generate realistic, context-rich images with GPT Image 2. Human annotators localise suspicious regions and provide artefact categories and textual explanations, covering both local visual defects and higher-level commonsense or physical inconsistencies. The comparison with existing benchmarks highlights SafeIMG's emphasis on safety relevance, complex visual content and fine-grained diagnostic annotations.}
    \label{fig:teaser}
\end{figure}

Detecting AI-generated images has therefore become an important problem in multimodal safety. In the upper part of Figure \ref{fig:teaser}, existing studies have proposed benchmarks and detection methods for synthetic-image recognition~\citep{wang2020cnn,ojha2023towards,sha2022defake,zhu2023genimage}. One line of work focuses on general natural images, emphasizing large-scale data construction, cross-generator generalization, and robustness to common degradations~\citep{wang2020cnn,zhu2023genimage,park2024community}. Another line examines more specific domains, including scientific images, artistic images, originality judgment, and vision-language models (VLMs) for forged-content recognition~\citep{hu2026scifigdetect,li2025artwork,zhu2025animedl,chen2024gim}. These studies provide important foundations for understanding synthetic-image detection.

However, existing benchmarks remain insufficient for evaluating AI-generated images in safety-critical evidentiary scenarios. First, many classic benchmarks were constructed with earlier generators, such as BigGAN~\citep{brock2019large}, ADM~\citep{dhariwal2021diffusion}, GLIDE~\citep{nichol2021glide}, VQ-Diffusion~\citep{gu2022vector},  or early versions of Stable Diffusion~\citep{rombach2022high}. Although these generators were representative at the time, they do not capture the realism of recent image generation systems. Detectors that perform well on earlier synthetic images may therefore fail to generalize to images produced by current models~\citep{ojha2023towards,wang2023dire,yan2024sanity,park2024community}.
Second, even recent benchmarks \citep{zhu2023genimage,park2024community} primarily contain general-purpose or domain-specific imagery without an explicit focus on public and individual safety. It therefore remains unclear how reliably current detectors identify synthetic images across safety-sensitive contexts. Third, current evaluations often rely on overall detection accuracy, which provides limited diagnostic insight~\citep{zhu2023genimage,chen2024gim,yang2025xaigid}. In safety-oriented detection, image types may differ substantially in risk and difficulty. A detector may perform well on public-safety images but fail on transaction screenshots. It may detect news-like photographs but fail after cropping, re-encoding, screenshots, or social-media transformations. Without a fine-grained taxonomy of risk domains and evidence types, it is difficult to identify where detectors are reliable and where they remain unsafe.

A further limitation concerns the evidence used to make authenticity judgements. Some generated images contain visible local artefacts, such as malformed text, faces or hands. These cues may become less dependable as generation models improve. Visual realism also does not necessarily imply real-world plausibility. An image may appear photorealistic while containing implausible spatial relations, broken causal structure, commonsense conflicts or violations of physical constraints. Detecting these problems requires more than recognising low-level generator traces. It requires assessing whether the depicted content is coherent with the physical, logical and social context it claims to represent. Image-level labels cannot determine whether a detector has identified such anomalies or reached a correct decision for an unrelated reason.

To address these gaps, we introduce SafeIMG, a safety-oriented benchmark for AI-generated image detection. As shown in Figure~\ref{fig:teaser}, SafeIMG integrates three components: risk-oriented scenario design, structured image generation and fine-grained artefact annotation. Its scenario taxonomy comprises 12 categories across two domains, public safety and individual safety. The public-safety domain covers disasters, violence, transport accidents, hazardous incidents, infrastructure failures, crowd safety and public-health events. The individual-safety domain includes personal emergencies, transaction records, private communications, fabricated scene evidence and identity endorsement. Rather than treating these categories as generic visual classes, SafeIMG defines them according to their evidentiary functions and potential safety consequences.
For each category, a risk-aware scenario space specifies the relevant content, media forms and safety concerns. These specifications guide the construction of contextually grounded prompts, which are used to generate synthetic images with GPT Image 2. Human annotators subsequently localise suspicious regions, assign artefact categories and provide textual explanations for the identified anomalies. The annotation schema covers both explicit local defects and higher-level inconsistencies involving commonsense, scene logic and physical plausibility. Together, these components support evaluation of not only whether a detector recognises synthetic content, but also where and why it considers an image suspicious.

To thoroughly investigate the detection capabilities and characteristics of existing methods, we evaluate two major categories of AI-generated image detectors: specialized forensic models designed for synthetic-image detection (e.g., CNNSpot~\citep{wang2020cnn},
FreDect~\citep{frank2020leveraging}, LNP~\citep{liu2022detecting}), and VLMs with image understanding and forgery-recognition capabilities (e.g., GPT
series~\citep{openai2025gpt5,openai2026gpt55}, Claude series~\citep{anthropic2026claude47}, Qwen
Series~\citep{bai2025qwen3vl,qwen2026qwen35omni}). 

Across these evaluations, three consistent patterns emerge from the results. First, automatic systems remain markedly less sensitive than humans. The strongest VLM recognises 49.5\% of generated images, and the best specialised detector recognises 33.1\%, whereas human evaluators achieve 81.7\% accuracy. Performance also varies substantially across the 12 safety scenarios, indicating that aggregate scores conceal important category-specific failures. Second, model explanations cover only 29.8\% of human-annotated anomalies and focus predominantly on local cues involving text, faces and image texture. They are substantially less sensitive to commonsense conflicts and physical inconsistencies that require scene-level reasoning. 
Finally, robustness is also limited, with one representative model losing more than 90\% of its initial detection accuracy under severe dissemination-induced degradation. These findings show that safety-oriented image detection requires evaluation beyond aggregate accuracy, including scenario-specific reliability, explanatory validity and robustness under realistic propagation conditions.

Our contributions are summarized as follows:
\begin{itemize}[leftmargin=*]
    \vspace{-0.3cm}
    \item We formulate safety-oriented AI-generated image detection as an evaluation setting centred on public- and individual-safety contexts. This setting shifts the focus from generic authenticity classification towards risk-sensitive assessment of potentially misleading visual content.

    \item We introduce SafeIMG, a benchmark covering 12 safety-sensitive scenarios generated using GPT Image~2. Its risk-oriented taxonomy and scenario spaces specify the evidentiary function, relevant content, media forms and potential consequences of each category.

    \item We provide fine-grained human annotations that include suspicious-region localisation, artefact categories and textual explanations. The annotations cover both local visual defects and higher-level commonsense or physical inconsistencies, enabling diagnostic evaluation and human–AI explanation alignment analysis.

    \item We systematically evaluate multiple types of detection methods and find that current models more readily recognize explicit local defects but struggle to identify commonsense conflicts and physical anomalies, revealing a key capability gap in high-risk image detection.
\end{itemize}


\begin{figure}[t]
    \centering
    \includegraphics[width=17cm]{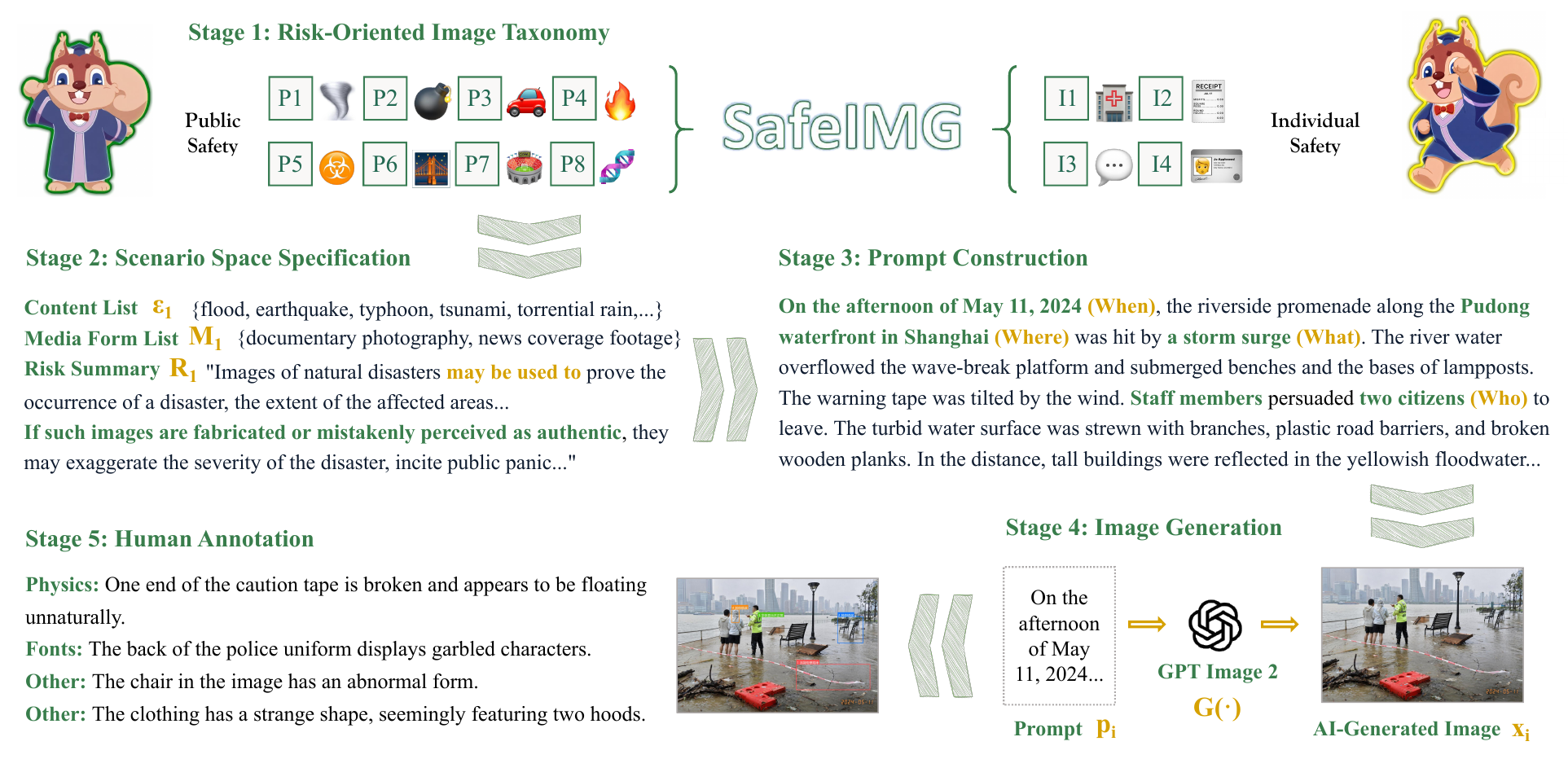}
    \vspace{-0.1cm}
    \caption{Construction pipeline of the SafeIMG benchmark for high-risk visual evidence scenarios. SafeIMG is built through five stages. 
    (1) The taxonomy in stage 1 defines 12 high-risk image categories under public or individual safety. 
    (2) For each category, the scenario space specifies the content list $\mathcal{E}$, media form list $\mathcal{M}$, and risk summary $R$; (3) complete prompts are then constructed before (4) generating images with GPT Image 2. 
    (5) Finally, human annotators localize suspicious regions, assign artifact labels, and provide textual explanations for fine-grained evaluation.}
    \label{fig:teaser-method}
        \vspace{-0.1cm}
\end{figure}

\vspace{-0.4cm}
\section{SafeIMG Benchmark}

SafeIMG is a safety-oriented benchmark of AI-generated images covering 12 scenarios across public- and individual-safety domains. The final benchmark contains two complementary subsets. The annotated subset comprises images with human-identified anomalies, each accompanied by suspicious-region localisation, a fine-grained artefact label and a textual explanation. Its annotations cover both conventional local defects and higher-level inconsistencies involving commonsense, scene logic and physical constraints. The extremely difficult subset contains highly realistic images for which annotators cannot identify a defensible visual anomaly. This design enables SafeIMG to evaluate both evidence-grounded detection and detection when no readily observable human cue is available. 

As shown in Figure~\ref{fig:teaser-method}, SafeIMG is constructed through five stages. First, the {Risk-Oriented Image Taxonomy} organises the benchmark into public- and individual-safety domains covering 12 high-risk scenarios. Second, {Scenario Space Specification} defines the relevant content and potential safety consequences of each category. Third, concrete scenario instances are converted into structured prompts. Fourth, GPT Image~2 generates a candidate pool of synthetic images. Finally, human screening removes invalid generations and separates the retained images into the annotated and extremely difficult subsets. Images entering the annotated subset are further labelled with suspicious regions, artefact categories and natural-language explanations.

\vspace{-0.2cm}
\subsection{Risk-Oriented Image Taxonomy}
\label{2.1}
Existing AI-generated image detection benchmarks are mostly organized based on general visual semantics, such as objects, styles, or generator sources. Such categorization is suitable for evaluating general image authenticity detection, but it is insufficient for covering safety issues in high-risk evidence scenarios. In real-world dissemination, the risk of an image depends not only on its visual content, but also on its evidentiary use.
Based on this consideration, to address the lack of safety-risk coverage and evidentiary image scenarios in existing benchmarks, SafeIMG constructs a risk-oriented image taxonomy. The data are organized according to the possible function of an image and its potential safety impact.
Specifically, as shown in Table \ref{tab:taxonomy}, SafeIMG divides high-risk images into two major groups: public safety and individual safety:

\textbf{\textit{The public safety group contains 8 categorie:}} P1 Natural Disasters; P2 Social Unrest, Public Violence, and Attack Incidents; P3 Transportation Accidents; P4 Fires, Explosions, and Energy Facility Accidents; P5 Pollution and Hazardous Material Accidents; P6 Building and Civil Infrastructure Accidents; P7 Crowd Gathering and Venue Safety Accidents; and P8 Public Health and Biosecurity Events. These images are often used to support claims about whether an event occurred, as well as the corresponding degree of damage or risk level. If such images are fabricated or misclassified as real, they may mislead the public's judgment of the authenticity and severity of the event, cause confusion or panic, and affect public opinion and social order.

\textbf{\textit{The individual safety group contains 4 categories:}} 
I1 Personal Accidents and Emergencies; I2 Private Receipts and Transaction Records; I3 Personal Chat and Communication Records; and I4 Fabricated Scene Evidence and Identity Endorsement. These categories focus on visual evidence that may be used in private communication, financial disputes, or personal identity claims. Compared with public safety scenarios, individual safety images are more directly related to the life and property safety of individuals. If such images are circulated or accepted as evidence, or even misused as tools for fraud, they may cause harm to personal property, reputation, privacy, and social trust.

Overall, the taxonomy of SafeIMG does not attempt to exhaust harmful forms of AI-generated images. Instead, it selects representative high-risk scenarios to provide a basis for benchmark construction. By organizing categories around evidentiary function, SafeIMG shifts the evaluation focus from general visual realism detection to risk-sensitive image authenticity detection. This taxonomy serves as the basis for subsequent scenario space specification, prompt construction, image generation, and human annotation.

\begin{table*}[t!]
\centering
\caption{Taxonomy of SafeIMG image categories and their evidentiary function.}
\renewcommand\arraystretch{1.1}
\setlength{\tabcolsep}{0.8mm}{
\scalebox{0.74}{
\begin{tabular}{p{0.03\textwidth} p{0.50\textwidth} p{0.78\textwidth}}
\toprule
\textbf{ID} & \textbf{Category} & \textbf{Evidentiary Function} \\
\midrule
\rowcolor{SafeGreen!25}
\multicolumn{3}{l}{\textbf{\textit{Public Safety Group}}} \\
P1 & Natural Disasters 
& Supports claims about disaster impact and emergency response. \\

P2 & Social Unrest, Public Violence, and Attack Incidents 
& Shapes claims about public order, security risks, and social stability. \\

P3 & Transportation Accidents 
& Supports claims about accident occurrence, damage, and responsibility. \\

P4 & Fires, Explosions, and Energy Facility Accidents 
& Supports risk assessment of hazardous incidents and emergency handling. \\

P5 & Pollution and Hazardous Material Accidents
& Supports claims that may trigger public panic or regulatory action. \\

P6 & Building and Civil Infrastructure Accidents 
& Supports claims about infrastructure safety, damage, and liability. \\

P7 & Crowd Gathering and Venue Safety Accidents 
& Supports judgments about crowd control, evacuation, and event safety. \\

P8 & Public Health and Biosecurity Events 
& Supports claims affecting public health decisions and social trust. \\
\midrule
\rowcolor{SafeGold!25}
\multicolumn{3}{l}{\textbf{\textit{Individual Safety Group}}} \\
I1 & Personal Accidents and Emergencies 
& Supports claims about personal danger, rescue needs, or liability. \\

I2 & Private Receipts and Transaction Records 
& Supports claims about payments, refunds, reimbursement, or disputes. \\

I3 & Personal Chat and Communication Records 
& Supports claims about private commitments, misconduct, or disputes. \\

I4 & Fabricated Scene Evidence and Identity Endorsement 
& Supports claims of presence, status, or endorsement. \\
\bottomrule
\end{tabular}
}}
\label{tab:taxonomy}
\end{table*}

\subsection{Scenario Space Specification}

In Section~\ref{2.1}, we define the risk categories. However, the content and risk focus vary greatly across categories. The taxonomy above only specifies which high-risk categories should be covered by the dataset, but it does not directly determine what should be generated for each category. Directly generating images from category names may lead to two problems. First, scenes within the same category may become repetitive and visually homogeneous. Second, the risk focus of different categories may remain unclear, causing generated images to present only generic visual content without a clear risk orientation. Therefore, to ensure the quality and diversity of synthetic images, we specify a generation scope for each category, namely a Scenario Space, to guide prompt construction under different categories.

Specifically, the Scenario Space of each category contains three components: (1) a risk summary $R_c$; (2) a content list $\mathcal{E}_c$; and (3) a media form list $\mathcal{M}_c$. The risk summary explains the core reason why images in this category may cause harm when fabricated or misused. The content list specifies which concrete events, objects, or materials can be generated under this category. The media form list specifies the visual forms in which such content commonly appears. Through this step, SafeIMG converts abstract risk categories into operable generation scopes.
\begin{equation}
    \mathcal{S}_c = \{(R_c, e, m) \mid e \in \mathcal{E}_c,\ m \in \mathcal{M}_c\}.
\end{equation}
For public safety images, the risk summary indicates the potential harm of images in the corresponding category. For example, the risk of P1 Natural Disasters lies in the possibility of exaggerating disaster severity, creating panic, or misleading rescue decisions. The content list mainly corresponds to different public events or accident scenes. For example, the content list of P1 includes 11 specific disaster types, such as floods, earthquakes, typhoons, and heavy rain. The media form list of public safety images mainly includes documentary photography and news-report images.

For individual safety images, the risk summary indicates the specific personal-rights scenarios in which such images may be involved. For example, images in I2 Private Receipts and Transaction Records may be used in telecom fraud. The content list may include either concrete events or various types of personal evidentiary materials. For example, Private Receipts and Transaction Records mainly include payment screenshots, refund records, transfer vouchers, and order pages. The media forms of individual safety images are more diverse, mainly including documentary photography, electronic-device screenshots, and photos taken of mobile phone screens.

Through Scenario Space Specification, SafeIMG establishes an intermediate layer between taxonomy and prompt construction. The taxonomy determines which risk categories the dataset covers. The scenario space specifies the generation scope: what content can be generated under each category, what media forms can be used, and what risk focus should be highlighted. Subsequent prompt construction can then build complete prompts based on this scenario space. Scenario Space Specification can reduce repetitive generation within the same category and improve the relevance of synthetic images to high-risk scenarios. This stage does not directly generate final prompts. Instead, it specifies the generation scope of each category and lays the foundation for subsequent sample-level prompt construction.

\subsection{Prompt Construction}

In the previous section, we define the scenario space $\mathcal{S}_c$ for each risk category, specifying the content scope, media forms, and risk focus of that category. This section further constructs complete prompts based on the scenario space of each category. Specifically, we synthesize complete prompts in batches for each category. For each sample under a specific category, we first randomly select a concrete scenario instance from the scenario space of that category, and then expand it into a complete generation instruction using universal prompt expansion rules and category-specific constraints. Formally, this process can be written as:
\begin{equation}
    p_i = \Psi(U, s_i, A_c), \quad s_i \in \mathcal{S}_c.
\end{equation}
Here, $p_i$ denotes the final prompt. $\Psi(\cdot)$ denotes the prompt constructor, $U$ denotes the universal prompt expansion rules, $s_i$ denotes a concrete scenario instance, and $A_c$ is the category-specific constraints, specificity:

\textbf{\textit{The prompt expansion rules $U$}} are shared by all categories and control the overall narrative structure of the prompt. We require the model to use the \textbf{5W+1H} structure commonly used in news narration, namely Who, What, When, Where, Why, and How. This structure ensures that the prompt has basic scenario completeness, so that the generated result contains a relatively clear event background and detailed information. For public safety images, 5W+1H helps supplement the location, time, cause, affected subjects, and on-site process of the event. For individual safety images, this narrative structure also makes the background more detailed and the evidentiary chain more reliable. In this way, 5W+1H serves not only as a descriptive template, but also as a constraint on scenario completeness, making the generated images more concrete and credible.

\textbf{\textit{The concrete scenario instance $s_i$}} is selected from the scenario space of a specific category. It usually determines the basic content framework of the sample, including the main event or material type, the adopted media form, and the main risk orientation. For example, in the Natural Disasters category, a scenario instance can be ``a news-report image showing severe urban road flooding caused by heavy rain, with high dissemination potential.''

\textbf{\textit{The category-specific constraints $A_c$}} supplements the special requirements of different categories in prompt construction. These constraints do not change the main content of the scenario instance, but they specify forms or details that are important for credibility. Different categories have different category-specific constraints. For example, for categories under the public safety group, we require the image to contain elements that indicate the time of the event. For I2 and I3 under the individual safety group, we require the model to explicitly avoid obvious placeholders such as ``xx'' or ``1234'' in common textual elements, including text, ID numbers, names, amounts, timestamps, and institutional identifiers. Instead, the generated content should be more natural, random, and contextually appropriate. The complete category-specific constraint settings are provided in the appendix.

In practice, we combine $U$, $s_i$, and $A_c$ to convert abstract categories into implementable sample-level prompts. Through this process, SafeIMG improves prompt specificity, scenario diversity, and risk relevance while maintaining category consistency, thereby laying the foundation for subsequent image generation.

\subsection{Image Generation}

After constructing the prompts, SafeIMG uses GPT Image 2 to generate synthetic images for the defined safety scenarios. Given a complete prompt $p_i$, the generation process is written as
\begin{equation}
    x_i = G(p_i),
\end{equation}
where $G$ denotes the image-generation model and $x_i$ is the resulting image.

To preserve the provenance of each image, we associate $x_i$ with its safety category $c_i$, scenario instance $s_i$, generation prompt $p_i$ and media form $m_i$. We define the resulting sample record as
\begin{equation}
    z_i = (x_i,c_i,s_i,p_i,m_i).
\end{equation}
The complete pool of generated candidates is then
\begin{equation}
    \mathcal{D}_{\mathrm{gen}}
    =
    \left\{z_i\right\}_{i=1}^{N_{\mathrm{gen}}},
\end{equation}
where $N_{\mathrm{gen}}$ is the number of images produced before human screening. At this stage, $\mathcal{D}_{\mathrm{gen}}$ contains all generated candidates, including valid images, extremely difficult images and invalid generations. Each sample can be traced to its category, scenario specification and generation instruction.

\subsection{Human Annotation}
\label{sec:human_annotation}

Most AI-generated image detection datasets provide only image-level authenticity labels. Such labels indicate whether an image is synthetic but do not reveal the evidence supporting that judgement. SafeIMG therefore introduces evidence-level human annotation to record where an image appears suspicious, what type of anomaly is present and why the anomaly is implausible.

A central feature of our annotation design is its scope. SafeIMG does not restrict annotation to familiar local defects, such as malformed text, faces, hands or lighting. Annotators are also instructed to examine higher-level inconsistencies that require an understanding of the depicted scene and the real world. These include implausible spatial relations, inconsistent event logic, violations of commonsense and conflicts with physical constraints. This distinction enables SafeIMG to assess whether detection models move beyond superficial generation artefacts towards evaluating the overall plausibility of visual content. The annotation protocol comprises two stages: sample triage and evidence-level annotation.

\textbf{Stage 1: Sample triage.}
Annotators first screen every sample $z_i\in\mathcal{D}_{\mathrm{gen}}$ and assign a triage label:
\begin{equation}
    q_i = \tau(z_i)
    \in
    \left\{
    \mathrm{ann},
    \mathrm{hard},
    \mathrm{invalid}
    \right\},
\end{equation}
where $\tau(\cdot)$ denotes the human-screening procedure.

Samples are labelled $\mathrm{invalid}$ when they substantially deviate from the intended category, omit the principal subject, contain obvious placeholder text or exhibit severe generation failures. These samples are excluded from subsequent annotation and evaluation. The discarded set is defined as:
\begin{equation}
    \mathcal{D}_{\mathrm{invalid}}
    =
    \left\{
    z_i\in\mathcal{D}_{\mathrm{gen}}
    \,\middle|\,
    q_i=\mathrm{invalid}
    \right\}.
\end{equation}

Samples are labelled $\mathrm{hard}$ when they are visually convincing and annotators cannot identify a defensible anomaly or suspicious region. These samples form the extremely difficult subset:
\begin{equation}
    \mathcal{D}_{\mathrm{hard}}
    =
    \left\{
    \left(z_i,\varnothing\right)
    \,\middle|\,
    z_i\in\mathcal{D}_{\mathrm{gen}},
    q_i=\mathrm{hard}
    \right\},
\end{equation}
where $\varnothing$ indicates that no defensible region-level anomaly was identified by the annotators.

The remaining valid images contain at least one identifiable anomaly and are assigned the label $\mathrm{ann}$. These images proceed to evidence-level annotation. 
This triage distinguishes poor-quality generation failures from highly realistic images without readily observable defects. It also ensures that the extremely difficult samples remain part of SafeIMG rather than being incorrectly discarded as annotation failures.

\textbf{Stage 2: Evidence-level annotation.}
For each sample assigned to the annotated branch, annotators examine both local appearance defects and the consistency of the overall scene. The annotation taxonomy contains two complementary groups. Local artefacts include abnormalities involving text, faces, hands, lighting and other directly observable visual details. High-level anomalies include commonsense conflicts, implausible object interactions, inconsistent event logic and violations of physical constraints.

Each identified anomaly is documented using three components. Annotators first draw a bounding box around the relevant object or suspicious region. They then assign a fine-grained artefact label describing the anomaly type. Finally, they provide a concise natural-language explanation of why the selected content appears abnormal or incompatible with the depicted scene.

For high-level anomalies, the explanation specifies the violated commonsense expectation, logical relation or physical constraint. An image may contain multiple independently annotated anomalies. The annotations associated with image $x_i$ are represented as
\begin{equation}
    \mathcal{A}_i
    =
    \left\{
    a_{ij}
    \right\}_{j=1}^{K_i},
    \qquad
    a_{ij}
    =
    \left(
    b_{ij},
    \ell_{ij},
    e_{ij}
    \right),
\end{equation}
where $K_i$ is the number of annotated anomalies. Here, $b_{ij}$ denotes the bounding box, $\ell_{ij}$ the artefact category and $e_{ij}$ the textual explanation.
Finally, the annotated subset is defined as:
\begin{equation}
    \mathcal{D}_{\mathrm{ann}}
    =
    \left\{
    \left(z_i,\mathcal{A}_i\right)
    \,\middle|\,
    z_i\in\mathcal{D}_{\mathrm{gen}},
    q_i=\mathrm{ann},
    K_i\geq 1
    \right\}.
\end{equation}
This subset contains images with human-identified local or high-level anomalies and their corresponding evidence-level annotations.

The final SafeIMG benchmark is the disjoint union of these two subsets:
\begin{equation}
    \mathcal{D}_{\mathrm{SafeIMG}}
    =
    \mathcal{D}_{\mathrm{ann}}
    \sqcup
    \mathcal{D}_{\mathrm{hard}},
\end{equation}
where $\sqcup$ denotes a disjoint union. 

The two subsets support complementary evaluation objectives. $\mathcal{D}_{\mathrm{ann}}$ enables evidence-level analysis of local artefacts and commonsense conflicts. By contrast, $\mathcal{D}_{\mathrm{hard}}$ evaluates detectors on highly realistic images for which human observers cannot specify a reliable visual anomaly. Together, they allow SafeIMG to assess both explainable detection and detection under the absence of readily observable human evidence.

\section{Empirical Studies}
\subsection{Experimental Setup}
This section introduces the experimental setup on SafeIMG, including the evaluation data, evaluated models, and evaluation metrics. Our goal is to systematically evaluate the detection capability of existing methods in high-risk evidentiary image scenarios, and to further analyze whether the detection rationales produced by models are consistent with the visual issues annotated by humans.

\begin{figure*}[t]
    \centering
   \makebox[\textwidth][l]{%
    \begin{subfigure}[t]{0.45\textwidth}
        \centering
        \includegraphics[width=\linewidth]{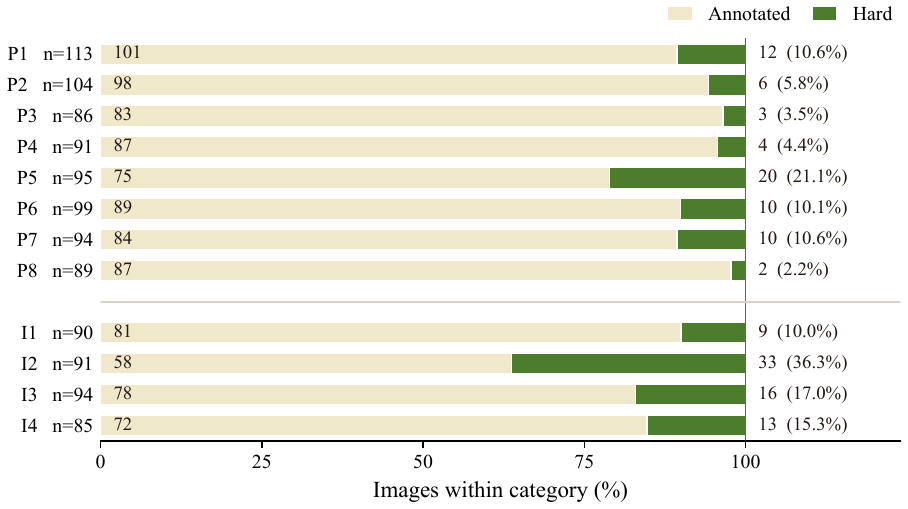}
        \caption{Image composition by safety category.}
        \label{fig:safeimg_image_composition}
    \end{subfigure}%
    \hspace{0.06\textwidth}%
    \begin{subfigure}[t]{0.42\textwidth}
        \centering
        \includegraphics[width=\linewidth]{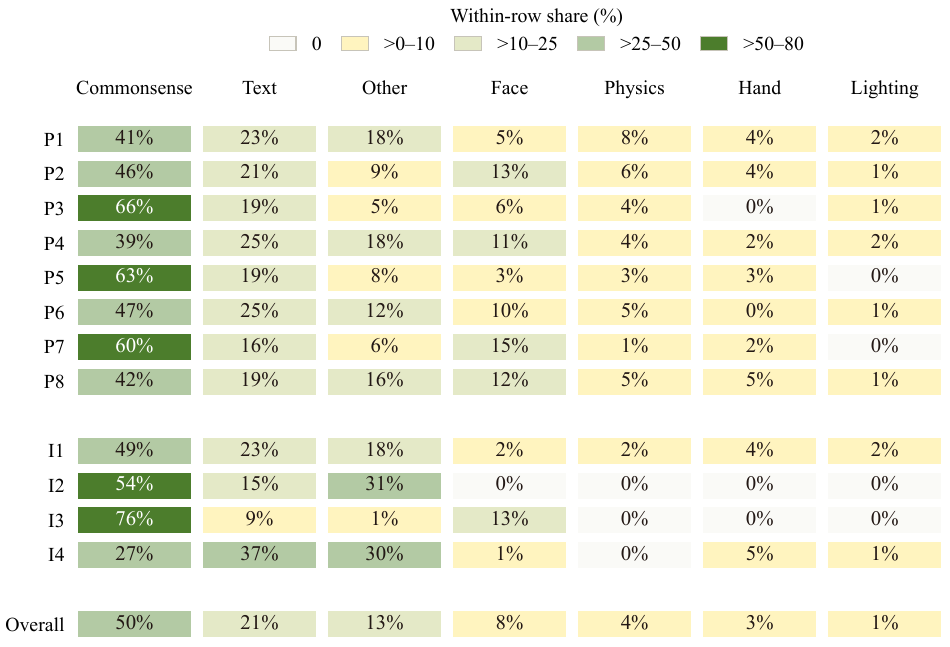}
        \caption{Anomaly-label composition by safety category.}
        \label{fig:safeimg_anomaly_evidence}
    \end{subfigure}%
}

    \vspace{0.35cm}

    \begin{subfigure}[t]{0.38\textwidth}
        \centering
        \includegraphics[width=\linewidth]{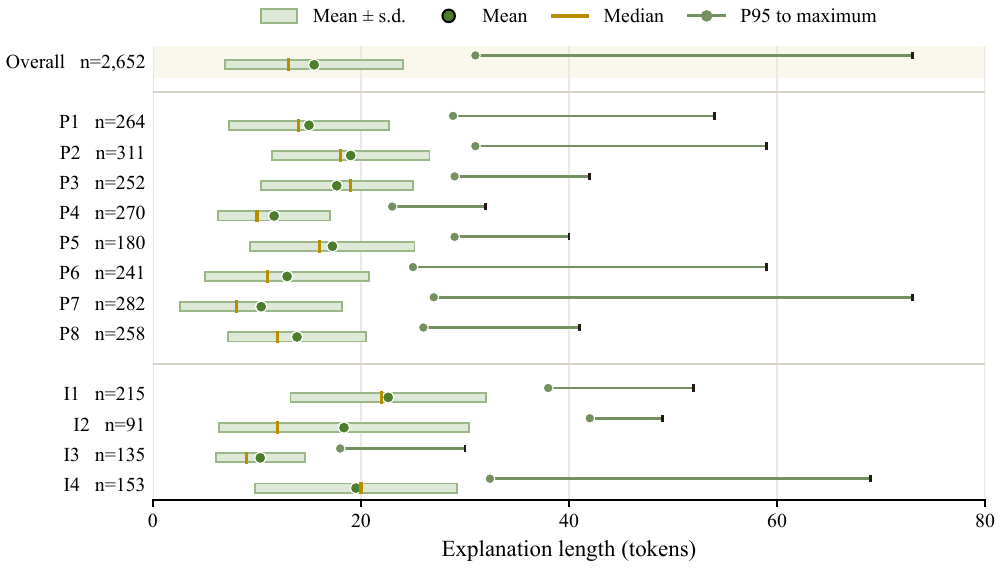}
        \caption{Explanation length by safety category.}
        \label{fig:safeimg_explanation_category}
    \end{subfigure}
    \hfill
    \begin{subfigure}[t]{0.58\textwidth}
        \centering
        \includegraphics[width=\linewidth]{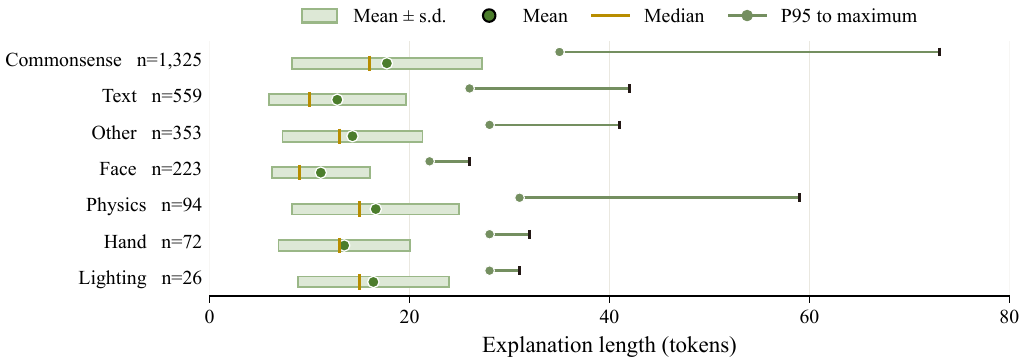}
        \caption{Explanation length by anomaly type.}
        \label{fig:safeimg_explanation_anomaly}
    \end{subfigure}
    \caption{SafeIMG evaluation data. (a) Category composition of the 1,131 synthetic images. Bars show the percentages assigned to the annotated (pale gold) and hard (green) subsets. Values give the corresponding image counts and hard-image rates. (b) Within-category distributions of the seven region-level anomaly labels. Cell values are rounded percentages, unrounded rows sum to 100\%. (c,d) Explanation-length summaries by safety category and anomaly type, respectively. Rectangles show the mean $\pm$ s.d., circles mark the means and vertical lines mark the medians. Horizontal segments span the 95th percentile to the maximum. P1--P8 denote public-safety categories and I1--I4 denote individual-safety categories.}
    \label{fig:safeimg_evaluation_data}
\end{figure*}

\textbf{Evaluation data.}
SafeIMG comprises 1,131 synthetic images generated with GPT Image~2 and spanning 12 safety-sensitive categories: eight public-safety categories and four individual-safety categories (Fig.~\ref{fig:safeimg_image_composition}). Human triage assigned 993 images (87.8\%) to the annotated subset $\mathcal{D}_{\mathrm{ann}}$ and 138 (12.2\%) to the hard subset $\mathcal{D}_{\mathrm{hard}}$. The hard-image fraction ranged from 2.2\% in P8 to 36.3\% in I2, revealing pronounced category-level variation in the visibility of local artefacts.

The annotated subset yielded 2,652 region-level annotations, equivalent to 2.67 regions per annotated image and 2.34 regions per benchmark image. Category-level means ranged from 1.57 (I2) to 3.36 (P7) regions per annotated image and from 1.00 to 3.00 regions per benchmark image. Each annotation includes a bounding box, one of seven anomaly labels and a concise English explanation. Commonsense violations were most frequent (50\%), followed by text artefacts (21\%) and other local defects (13\%). Faces (8\%), physical constraints (4\%), hands (3\%) and lighting (1\%) accounted for the remaining annotations (Fig.~\ref{fig:safeimg_anomaly_evidence}). Explanations contained 15.50 tokens on average (median, 13; s.d., 8.58; 95th percentile, 31; maximum, 73). Category- and anomaly-specific distributions are summarised in Figs.~\ref{fig:safeimg_explanation_category} and~\ref{fig:safeimg_explanation_anomaly}, respectively.

We use $\mathcal{D}_{\mathrm{ann}}$ to quantify artefact prevalence, model coverage of human-identified anomalies and human--AI explanation alignment. We evaluate $\mathcal{D}_{\mathrm{hard}}$ separately to test whether models recognise synthetic images when no defensible local anomaly can be identified.

To estimate class-specific prediction bias, we further construct $\mathcal{D}_{\mathrm{real}}$, a balanced reference set of 600 genuine images. The set contains 50 images in each of the 12 safety categories and spans source material dated from 2000 to 2026. All public-safety images (P1--P8) were sourced from Wikimedia Commons. For the individual-safety categories, I1 came from MEDIC~\citep{alam2023medic}, I2 from WildReceipt~\citep{sun2021spatial}, and I3 from RICO~\citep{deka2017rico}. I4 were acquired through internet. It includes 348 landscape, 190 portrait and 62 square images, with a mean resolution of 1.287 megapixels.

\textbf{Evaluated Models.}
We evaluate two types of AI-generated image detection methods. The first type is general-purpose vision-language models (VLMs), including multimodal foundation models from GPT, Gemini, Claude, Qwen, Doubao, and others. These models possess strong visual understanding and natural-language explanation capabilities, and thus can be directly used for authenticity judgment and rationale generation. For these models, we use a unified prompt requiring the model to output an image label, a confidence score, and several independent visual rationales. The output label is restricted to either \texttt{real} or \texttt{ai\_generated}; the confidence score is represented by discrete levels; and each rationale is required to be an independently checkable visual observation.

The second type is specialized AI-generated image detection models. These methods usually directly output whether an image is AI-generated, or produce a detection score related to the probability of AI generation. For detectors that provide continuous scores, we convert them into binary labels according to their default thresholds or officially recommended settings, while preserving the original scores for subsequent analysis. Unlike general-purpose VLMs, specialized detectors usually do not provide natural-language explanations, and are therefore mainly used for image-level detection performance comparison.

In addition to evaluating these two types of models, we conducted a human evaluation. We invited three human evaluators to independently perform the authenticity assessment without consulting one another. These evaluators were independent of the annotators who conducted the evidence-level annotation described in Section~\ref{sec:human_annotation}. We calculated the evaluation metric separately for each evaluator and report the arithmetic mean across the three evaluators.

\textbf{Evaluation metrics.}
SafeIMG evaluates detection models at both the image and evidence levels. Unless otherwise specified, generated-image detection is evaluated on the complete benchmark $\mathcal{D}_{\mathrm{SafeIMG}}$. Fine-grained artefact and explanation analyses are restricted to $\mathcal{D}_{\mathrm{ann}}$, because $\mathcal{D}_{\mathrm{hard}}$ does not contain region-level annotations. We additionally report performance on $\mathcal{D}_{\mathrm{hard}}$ to assess detection when no readily observable human cue is available.
For a model $M$ and a generated-image set $\mathcal{D}$, we define the AI detection rate by
\begin{equation}
    \frac{1}{|\mathcal{D}|}
    \sum_{x_i\in\mathcal{D}}
    \mathbb{I}
    \left[
    \hat{y}_{M}(x_i)=\texttt{ai\_generated}
    \right],
\end{equation}
where $\hat{y}_{M}(x_i)$ denotes the predicted label and $\mathbb{I}[\cdot]$ is the indicator function. Because every image in $\mathcal{D}$ is synthetic, the AI rate is equivalent to recall for the AI-generated class. It measures generated-image sensitivity and should not be interpreted as complete binary-classification accuracy.




\subsection{Overall Detection Performance}

We first report the overall image-level detection performance of different models on SafeIMG. This section provides an integrated overview of the detection capability of existing methods, including general-purpose VLMs, specialized detectors, and human evaluators, within the context of high-risk evidentiary image scenarios. The overall results, detailed in Table~\ref{tab:vlm_accuracy}, reveal that current models and methods face clear limitations when identifying risk-oriented images synthesized by frontier generative models.

At a high level, the best-performing general-purpose VLM achieves a detection accuracy of 49.5\%, while the best-performing specialized image detector reaches only 33.1\%. In stark contrast, human evaluators attain a significantly higher average accuracy of 81.7\%, underscoring the substantial gap between current automated systems and human-level forensic judgment.

Different VLMs exhibit substantial performance differences, with some models showing particularly poor discriminative ability for AI-generated images. This indicates that general visual understanding ability does not naturally translate into reliable AI-generated image detection capability. Furthermore, stronger model capability or more intensive reasoning settings do not always lead to consistent improvements. For instance, within the GPT series, the no-think setting occasionally outperforms the high-reasoning setting on several public safety categories. This suggests that VLM detection capability is highly model-dependent and cannot be simply predicted by scale or reasoning intensity.

\begin{table*}[t!]
\centering
\caption{Detection accuracy (\%) of VLMs, specialized detectors, and human evaluators on SafeIMG. Overall is weighted by category sample size. Within each model group, the highest value in each column is highlighted in blue and the second-highest value is highlighted in pink.}
\label{tab:vlm_accuracy}
\renewcommand\arraystretch{1.1}
\setlength{\tabcolsep}{1.35mm}
\scriptsize
\resizebox{\textwidth}{!}{
\begin{tabular}{
>{\raggedright\arraybackslash}p{40mm}
ccccccccccccc
}
\toprule
\rowcolor{SafeGreen!70}
\textbf{Model} &
\textbf{P1} & \textbf{P2} & \textbf{P3} & \textbf{P4} &
\textbf{P5} & \textbf{P6} & \textbf{P7} & \textbf{P8} &
\textbf{I1} & \textbf{I2} & \textbf{I3} & \textbf{I4} &
\textbf{Overall} \\
\midrule

\rowcolor{SafeGold!25}
\multicolumn{14}{l}{\textbf{\textit{VLM Base Models}}} \\

\textbf{Claude Opus 4.6}
& 30.4 & 42.3 & 29.1 & 29.7 & 33.0 & 16.2 & 17.0
& \best{46.1} & 12.2 & \best{53.8} & \best{78.7} & 25.9 & 32.6 \\

\textbf{Claude Sonnet 4.6}
& 20.5 & 30.8 & 29.1 & 22.0 & 24.5 & 11.1 & 12.8
& 22.5 & 4.4 & \secondbest{42.9} & \secondbest{46.8} & 20.0 & 21.7 \\

\textbf{GPT 5.5 (reason effort = high)}
& 6.2 & 13.5 & 23.3 & 12.1 & 12.6 & 4.0 & 4.3
& 4.5 & 1.1 & 11.0 & 8.5 & 0.0 & 8.9 \\

\textbf{GPT 5.4 (reason effort = high)}
& 0.0 & 1.0 & 0.0 & 1.1 & 1.1 & 0.0 & 0.0
& 0.0 & 0.0 & 4.4 & 1.1 & 0.0 & 0.5 \\

\textbf{GPT 5.5 (no think)}
& 17.7 & 22.1 & 32.6 & 18.7 & 26.3 & 13.1 & 6.4
& 21.3 & 1.1 & 5.5 & 7.4 & 2.4 & 15.3 \\

\textbf{GPT 5.4 (no think)}
& 3.5 & 17.3 & 15.1 & 5.5 & 6.3 & 4.0 & 4.3
& 4.5 & 0.0 & 3.3 & 4.3 & 0.0 & 6.2 \\

\textbf{Doubao Seed 2.0 Pro}
& 38.1 & \secondbest{52.9} & 47.7 & 38.5 & \secondbest{41.1}
& 22.2 & 45.7 & 39.3 & 24.4 & 34.1 & 40.4
& \best{42.4} & 38.2 \\

\textbf{Doubao Seed 2.0 Mini}
& \best{54.4} & \best{63.2} & \best{57.0} & \best{64.8}
& \best{60.0} & \best{54.5} & \best{57.4} & \secondbest{42.7}
& \best{53.1} & 21.4 & 40.7 & \secondbest{40.8} & \best{49.5} \\

\textbf{Gemini 3.1 Pro}
& \secondbest{43.4} & 48.1 & \secondbest{54.7} & \secondbest{46.2}
& 31.6 & \secondbest{38.4} & \secondbest{52.1} & 40.4
& \secondbest{35.6} & 20.9 & 26.6 & 37.6 & \secondbest{39.5} \\

\textbf{Qwen3.6 Flash}
& 24.3 & 21.4 & 32.6 & 22.0 & 16.0 & 16.2 & 14.9
& 13.5 & 4.4 & 21.1 & 19.4 & 10.8 & 18.1 \\

\textbf{Qwen3.6 Plus}
& 23.9 & 27.9 & 37.2 & 28.6 & 26.3 & 20.2 & 20.2
& 16.9 & 4.4 & 24.2 & 27.7 & 23.5 & 24.1 \\

\midrule
\rowcolor{SafeGold!25}
\multicolumn{14}{l}{\textbf{\textit{Specialized AI Detection Models}}} \\

\textbf{CNNSpot}
& 9.7 & 6.7 & 8.1 & \secondbest{17.6} & \secondbest{12.6}
& 11.1 & 1.1 & 2.2 & \secondbest{2.2} & \secondbest{4.4}
& 0.0 & 0.0 & 7.3 \\

\textbf{FreDect}
& \best{33.6} & \best{20.2} & \best{41.9} & \best{42.9}
& \best{37.9} & \best{41.4} & \best{18.1} & \best{33.7}
& \best{22.2} & \best{45.1} & \best{38.3} & \best{41.2}
& \best{33.1} \\

\textbf{LNP}
& \secondbest{13.3} & \secondbest{9.6} & \secondbest{18.6}
& 7.7 & \secondbest{12.6} & \secondbest{14.1}
& \secondbest{8.5} & \secondbest{14.6} & \secondbest{2.2}
& 0.0 & \secondbest{1.1} & \secondbest{1.2}
& \secondbest{9.1} \\

\midrule
\rowcolor{SafeGold!25}
\multicolumn{14}{l}{\textbf{\textit{Human Evaluation}}} \\

\textbf{Human}
& \best{83.2} & \best{81.1} & \best{80.2} & \best{82.4}
& \best{77.5} & \best{78.8} & \best{82.3} & \best{86.9}
& \best{93.3} & \best{70.7} & \best{91.1} & \best{71.8}
& \best{81.7} \\

\bottomrule
\end{tabular}
}
\end{table*}

Existing specialized detectors show limited performance on SafeIMG due to a clear distribution shift problem. In particular, early methods like CNNSpot and LNP, which learned statistical artifacts , achieve only 7.3\% and 9.1\% accuracy, respectively, while FreDect performs relatively better at 33.1\%. This indicates that frequency-domain cues retain some value, but relying on low-level traces is insufficient for images produced by newer generators like GPT Image 2. The performance of these detectors is also highly imbalanced across categories, suggesting that different evidentiary scenarios present distinct visual cues that no single detector can comprehensively cover.

From a category-level perspective, the difficulty is not uniform. Text-intensive or interface-like images, such as private receipts (I2) and chat records (I3), are easier for some VLMs to identify due to rich textual and layout cues. In contrast, categories resembling real photographs or news scenes, such as personal accidents (I1), infrastructure accidents (P6), and crowd gatherings (P7), are more challenging for both VLMs and specialized detectors. This pattern is echoed in human evaluation, where complex backgrounds, occlusions, and crowds (e.g., in P6, P7, and P8) also increase the difficulty of manual recognition.

Overall, the differences in performance across scenarios such as public safety images, private receipts, chat records, and identity endorsement further demonstrate that SafeIMG is not merely a test of identifying generic visual defects. Instead, it poses a comprehensive challenge involving scenario understanding, text/interface consistency, commonsense reasoning, and forensic sensitivity. The results validate the necessity of SafeIMG as a safety-oriented benchmark, highlighting that current automated systems are insufficient for high-stakes environments and that manual review, while more accurate, is also prone to error and should be complemented by automatic detection, cross-modal consistency verification, and source-credibility analysis.

\begin{figure}[t]
    \centering
    \includegraphics[width=16cm]{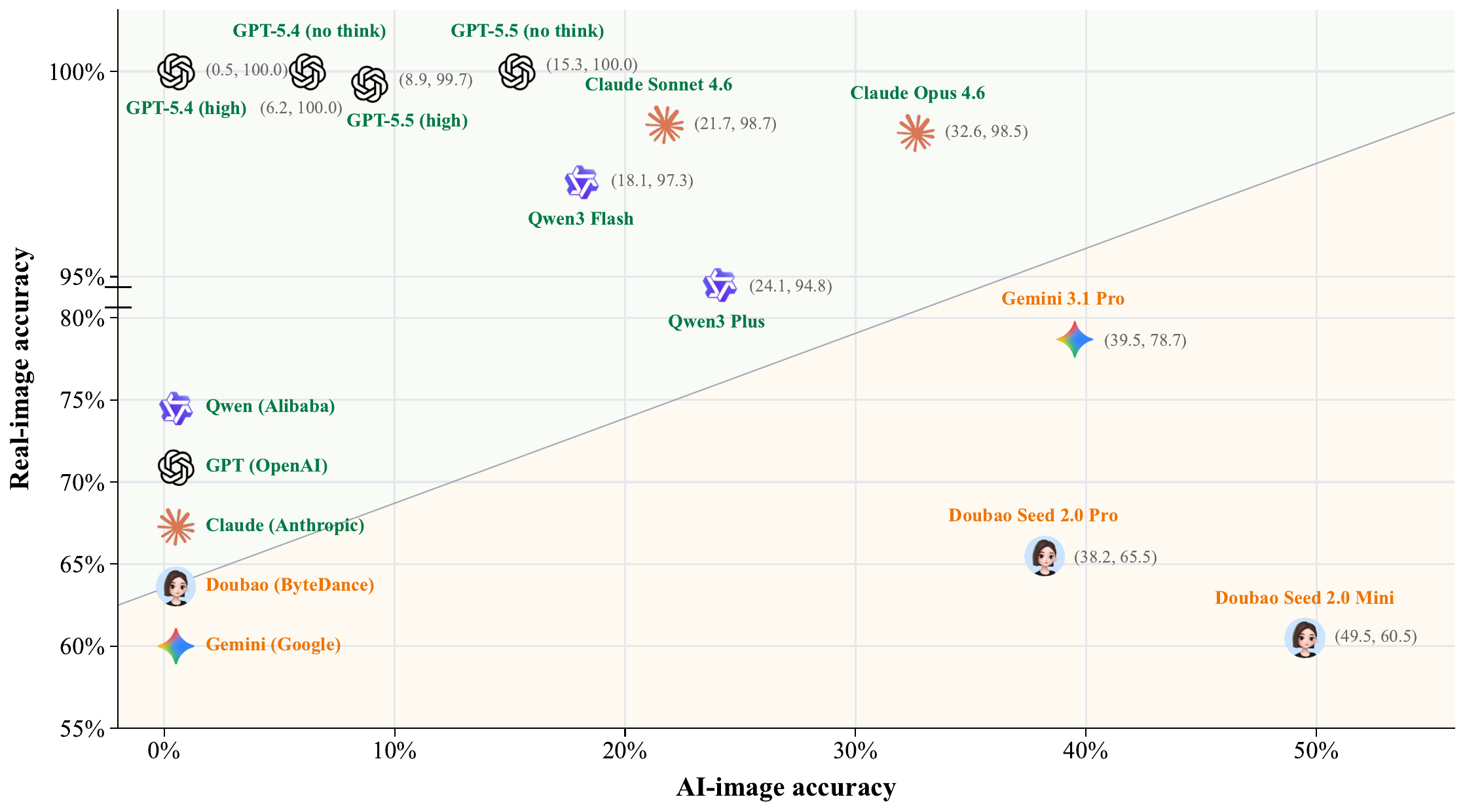}
    \caption{Trade-off between AI-image accuracy and real-image accuracy across VLMs on SafeIMG. Points above the diagonal indicate higher AI than real accuracy, and vice versa.}
    \label{fig:scatter}
\end{figure}

\subsection{Trade-off between AI and Real Image Accuracy}

Beyond overall detection accuracy, we further examine how different VLMs balance the identification of AI-generated images in $\mathcal{D}_{\mathrm{SafeIMG}}$ against the correct classification of real images in $\mathcal{D}_{\mathrm{real}}$. Figure~\ref{fig:scatter} plots each model's accuracy on AI-generated images against its accuracy on real images, revealing a clear trade-off that separates the models into two distinct groups.

The first group, comprising GPT-5.4, GPT-5.5, Claude Opus 4.6, Claude Sonnet 4.6, Qwen3 Flash, and Qwen3 Plus, exhibits a strong real-image bias. These models achieve extremely high real-image accuracy ranging from 94.8\% to 100.0\%, but their AI-image accuracy remains disappointingly low, with GPT-5.4 (reason effort = high) dropping to only 0.5\%. This indicates that such models are heavily conservative in making "AI-generated" judgments, tending to classify almost all inputs as real to avoid false positives. While this behavior is desirable in applications where misclassifying real images as fake carries high costs, it renders these models nearly ineffective for AI-generated image detection in safety-critical evidentiary scenarios.

The second group, including Doubao Seed 2.0 Mini, Doubao Seed 2.0 Pro, and Gemini 3.1 Pro, adopts a more balanced strategy. Doubao Seed 2.0 Mini achieves the highest AI-image accuracy among all VLMs at 49.5\%, while maintaining a moderate real-image accuracy of 60.5\%. Similarly, Doubao Seed 2.0 Pro and Gemini 3.1 Pro achieve AI-image accuracies of 38.2\% and 39.5\%, with real-image accuracies of 65.5\% and 78.7\%, respectively. These models demonstrate that improving AI-image detection does not necessarily require sacrificing real-image performance entirely, though a clear trade-off remains.

Notably, no model in our evaluation achieves superior performance on both dimensions simultaneously. The GPT, Claude, and Qwen series prioritize real-image fidelity at the expense of detection sensitivity, while the Doubao and Gemini series lean toward higher detection rates at the cost of more frequent false alarms on real images. This divergence suggests that current VLMs have not yet converged on an optimal strategy for evidentiary image forensics, and the choice of model should depend on the specific risk tolerance of the deployment scenario.

Overall, the above analysis reinforces our finding that general visual understanding ability does not naturally translate into reliable AI-generated image detection. The pronounced trade-off between AI and real image accuracy highlights a fundamental challenge for current VLMs, which struggle to distinguish highly realistic synthetic content from genuine evidence, particularly when high precision on real images is required.

\begin{figure}[t]
\captionsetup[subfigure]{justification=centering,singlelinecheck=false}
    \centering
    \begin{subfigure}[t]{0.50\linewidth}
        \centering
        \includegraphics[width=\linewidth]{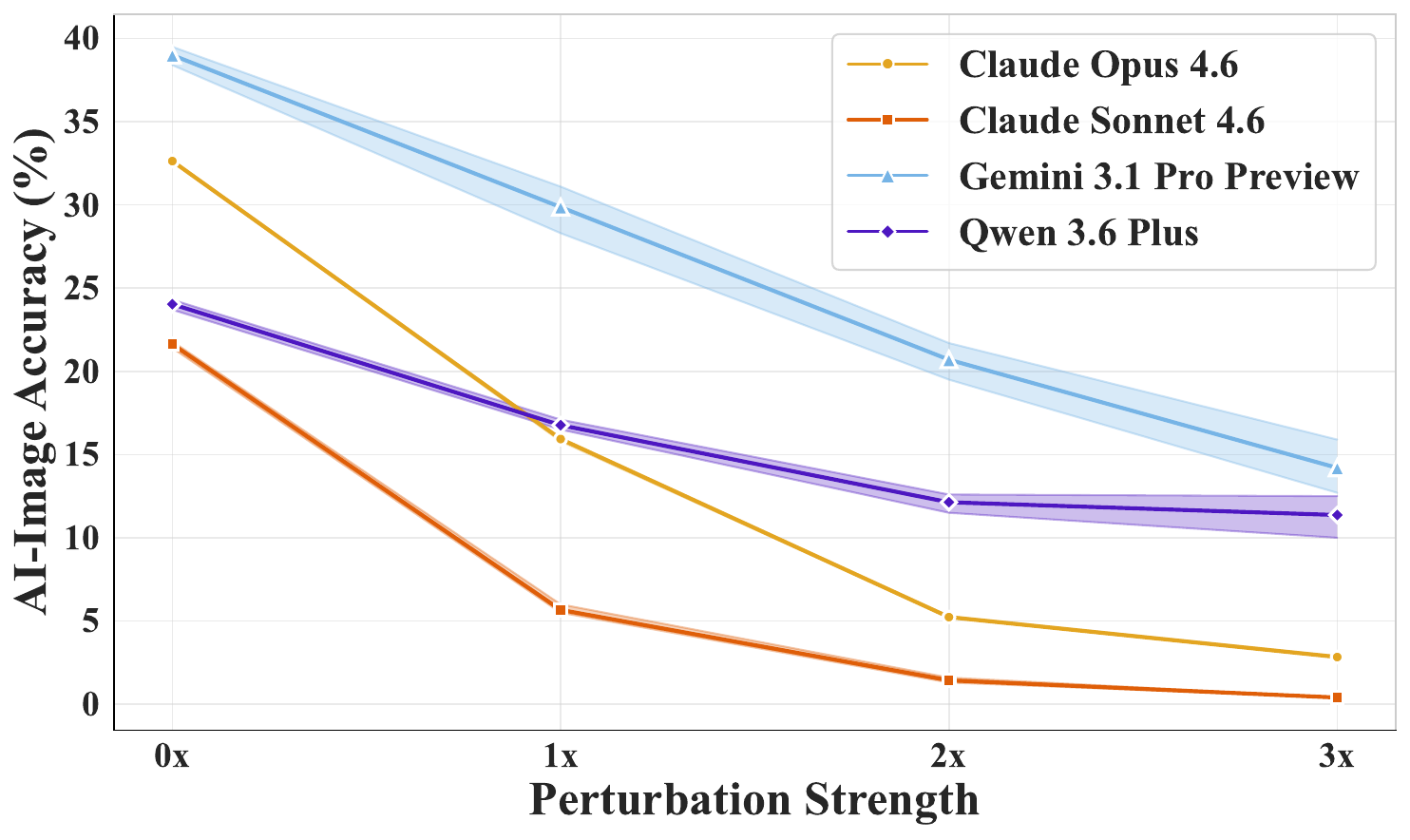}
        \caption{ }
        \label{fig:zhexian}
    \end{subfigure}%
    \hfill
    \begin{subfigure}[t]{0.43\linewidth}
        \centering
        \includegraphics[width=\linewidth]{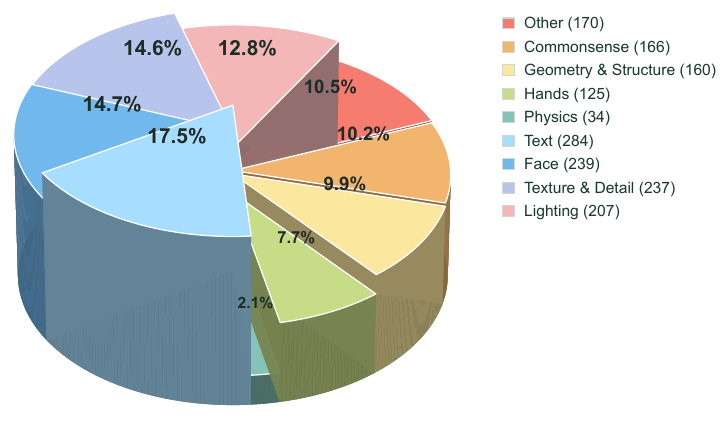}
        \caption{~~~~~~~~~~~~}
        \label{fig:humansupported}
    \end{subfigure}
    \caption{(a). AI-image detection accuracy of representative VLMs under network propagation perturbations. 
    (b). Distribution of model-generated rationales that are not supported by human annotations.}
    \label{fig:two_images}
\end{figure}

\subsection{Impact of Network Propagation on Detection Accuracy}

Beyond evaluating models on pristine AI-generated images, we further investigate how common network propagation perturbations affect detection performance in real-world social media scenarios. Figure~\ref{fig:zhexian} presents the detection accuracy of four VLMs under increasing levels of JPEG compression and image degradation, denoted as 0x, 1x, 2x, and 3x, where higher levels indicate more severe quality degradation.

Notably, all models exhibit a consistent downward trend in detection accuracy as perturbation intensity increases, confirming that network propagation poses a serious challenge to AI-generated image detection. However, the rate of decline varies substantially across models, revealing distinct robustness characteristics.
Gemini 3.1 Pro demonstrates the strongest overall robustness among the evaluated models, maintaining the highest accuracy across all perturbation levels. Its absolute performance remains superior throughout, though the rate of decline is moderate.
Claude Opus 4.6 exhibits the steepest relative drop among all models, falling from 32.6\% at 0x to merely 2.9\% at 3x, a reduction of over 90\% relative to its starting point. This sharp degradation implies that its detection capability relies heavily on high-frequency image details that are easily destroyed by compression. Qwen3 Plus, by comparison, shows the most gradual relative decline, decreasing from 24.1\% to 11.6\%, retaining nearly half of its initial accuracy even under severe perturbation. 
Claude Sonnet 4.6 performs the poorest across all settings, with accuracy plummeting to only 0.4\% at 3x, indicating that certain VLMs are almost entirely dependent on low-level visual artifacts, making them highly vulnerable to common image transformations encountered during social media sharing.

These results carry important practical implications. In real-world evidentiary scenarios, images are rarely preserved in their original quality, as they are frequently compressed, resized, and re-encoded during online dissemination. The substantial performance drop observed across all models underscores that current VLMs lack the robustness required for reliable deployment in forensic applications. Furthermore, the pronounced variation in degradation sensitivity across models suggests that robustness to network propagation should be considered a critical evaluation dimension in future benchmark development, alongside pristine-image detection accuracy.


\subsection{Human-AI Alignment in Artifact Explanations}
In the previous sections, we analyzed the performance of detection models on SafeIMG images. We next examine whether model-generated rationales align with the evidence identified by human annotators. Because this analysis requires fine-grained reference annotations, all experiments in this section are conducted on $\mathcal{D}_{\mathrm{ann}}$. We use a language model to compare the semantic content of human annotations with model-generated rationales and quantify their agreement. A match does not require identical wording, it is recorded when a model rationale identifies the same problem or anomaly described in a human annotation. AI-generated rationales evaluated in this experiment are produced by claude opus 4.6. And we use GPT-5.5 as the judge model for the Human–AI alignment analysis.

First, from the perspective of whether human-annotated artifacts are covered by the model, the alignment between the model and humans is still limited. For each image, we calculate the proportion of human-annotated artifacts that are hit by the model's reasons, and then average this score over the whole dataset. As shown in Table~\ref{tab:alignment}, Human Covered by AI is generally low across categories, with an average of only 29.8\%. This indicates that even when the model can make image-level authenticity judgments to some extent, its explanations still fail to sufficiently cover the key artifacts annotated by humans. Across categories, I4 and P4 have relatively higher Human Covered by AI scores, reaching 48.6\% and 49.9\%, respectively. This suggests that in fabricated evidence and identity endorsement scenarios, as well as fire and  explosion accidents scenarios, the model is relatively more likely to capture the anomalous cues that humans focus on. In contrast, I1, I2, and P5 have much lower coverage rates, at 19.1\%, 21.4\%, and 21.5\%, respectively. This indicates larger discrepancies between model explanations and human annotations in personal emergencies, private receipts and transaction records, and pollution or hazardous material incidents.

We further analyze why the consistency between model outputs and human annotations is low by examining which types of human-annotated artifacts are most often missed by the model. As shown in Figure~\ref{fig:doublepie}, 
the hit rate varies substantially across artifact types. The model achieves relatively higher coverage on Hands, Text, and Faces, with hit rates of 64.71\%, 58.67\%, and 56.25\%, respectively. These issues usually have clear local visual manifestations. In contrast, the hit rates for Commonsense and Physics are much lower, at only 15.01\% and 12.00\%, respectively. This shows that the model is more likely to capture local and explicit visual artifacts, while current models remain limited in recognizing problems that require higher-level understanding, such as scene logic, commonsense conflicts, and physical inconsistencies. In other words, the model's explanatory ability is still biased toward visible local defects and remains weaker in judging the overall plausibility of a scene.

\begin{table*}[t!]
\centering
\caption{Statistics of Human Covered by AI and AI Supported by Human across categories.}
\renewcommand\arraystretch{1.1}
\setlength{\tabcolsep}{1.5mm}{
\scalebox{0.9}{
\begin{tabular}{l c c c c c c c c c c c c c}
\toprule
\textbf{Metric} & \textbf{P1} & \textbf{P2} & \textbf{P3} & \textbf{P4} & \textbf{P5} & \textbf{P6} & \textbf{P7} & \textbf{P8} & \textbf{I1} & \textbf{I2} & \textbf{I3} & \textbf{I4} & \textbf{Overall} \\
\midrule
Human Covered by AI & 34.1 & 27.3 & 26.7 & 49.9 & 21.5 & 34.4 & 28.6 & 22.9 & 19.1 & 21.4 & 30.1 & 48.6 & 29.8 \\
AI Supported by Human & 18.5 & 17.6 & 11.4 & 21.2 & 8.6 & 17.9 & 18.1 & 16.7 & 6.9 & 5.8 & 9.3 & 24.8 & 14.3 \\
\bottomrule
\end{tabular}
}}
\label{tab:alignment}
\vspace{-0.2cm}
\end{table*}

In addition, we analyze the reasons proposed by the model that are not supported by human annotations. These reasons are not necessarily incorrect judgments or hallucinations. On the contrary, they may include key artifacts overlooked by humans, reflecting differences between humans and AI in what they focus on when judging images. Specifically, because the model's reasons do not come with fixed category labels, and because its focus does not fully match the human annotation taxonomy, we do not directly apply the human artifact labels to these reasons. Instead, we use a language model to conduct semantic summarization and keyword extraction, and finally group them into several thematic categories. As shown in Figure~\ref{fig:humansupported}, the model reasons not supported by human annotations are mainly concentrated in themes such as Text, Face, Texture\&Detail, Lighting, and Commonsense. Among them, Text accounts for the largest share, at 17.51\%; Face and Texture\&Detail account for 14.73\% and 14.61\%, respectively; Lighting accounts for 12.76\%; and Commonsense accounts for only 10.23\%.

These results show that when explaining images, the model tends to actively inspect cues such as text, human features, and texture details. This is consistent with our earlier findings: the model tends to focus on explicit and local texture or detail-level features in images, while there remains substantial room for improvement in higher-level logical issues that are truly important in human annotations, such as physical anomalies.

\subsection{Detection Accuracy on Challenging AI-Generated Images}
In this section, we further evaluate representative VLMs on $\mathcal{D}_{\mathrm{hard}}$. This subset contains highly realistic synthetic images for which human annotators could not identify a defensible visual anomaly. It therefore provides a challenging test of model accuracy when readily observable generation artefacts are absent.

As shown in Figure~\ref{fig:radar}, most models achieve substantially lower detection accuracy on $\mathcal{D}_{\mathrm{hard}}$ than on the complete SafeIMG benchmark. The accuracy of Doubao Seed 2.0 Pro decreases from 38.2\% to 22.5\%, corresponding to a relative reduction of 41.1\%. Similarly, the accuracy of Qwen3.6 Flash falls from 18.1\% to 10.1\%, a relative reduction of 44.2\%. These declines suggest that current detection capabilities depend partly on explicit visual cues that are unavailable in the hard subset. The subset therefore provides a more discriminative evaluation of highly realistic synthetic-image detection.

Detection accuracy also varies considerably across safety categories. No model maintains consistently strong performance across all scenarios. Gemini 3.1 Pro Preview performs relatively well on I1, I4 and several public-safety categories, whereas Claude Sonnet 4.6 achieves higher accuracy on I2 and I3. These fragmented strengths indicate that current detection capabilities remain category-dependent. Performance gains within one type of visual content do not consistently transfer to other public- or individual-safety scenarios. Overall, the results reveal limited generalisability when obvious generation artefacts are unavailable, with model judgements remaining sensitive to image content and scenario characteristics.

\subsection{Case Study}

\begin{figure}[t]
    \centering
    \includegraphics[width=14cm]{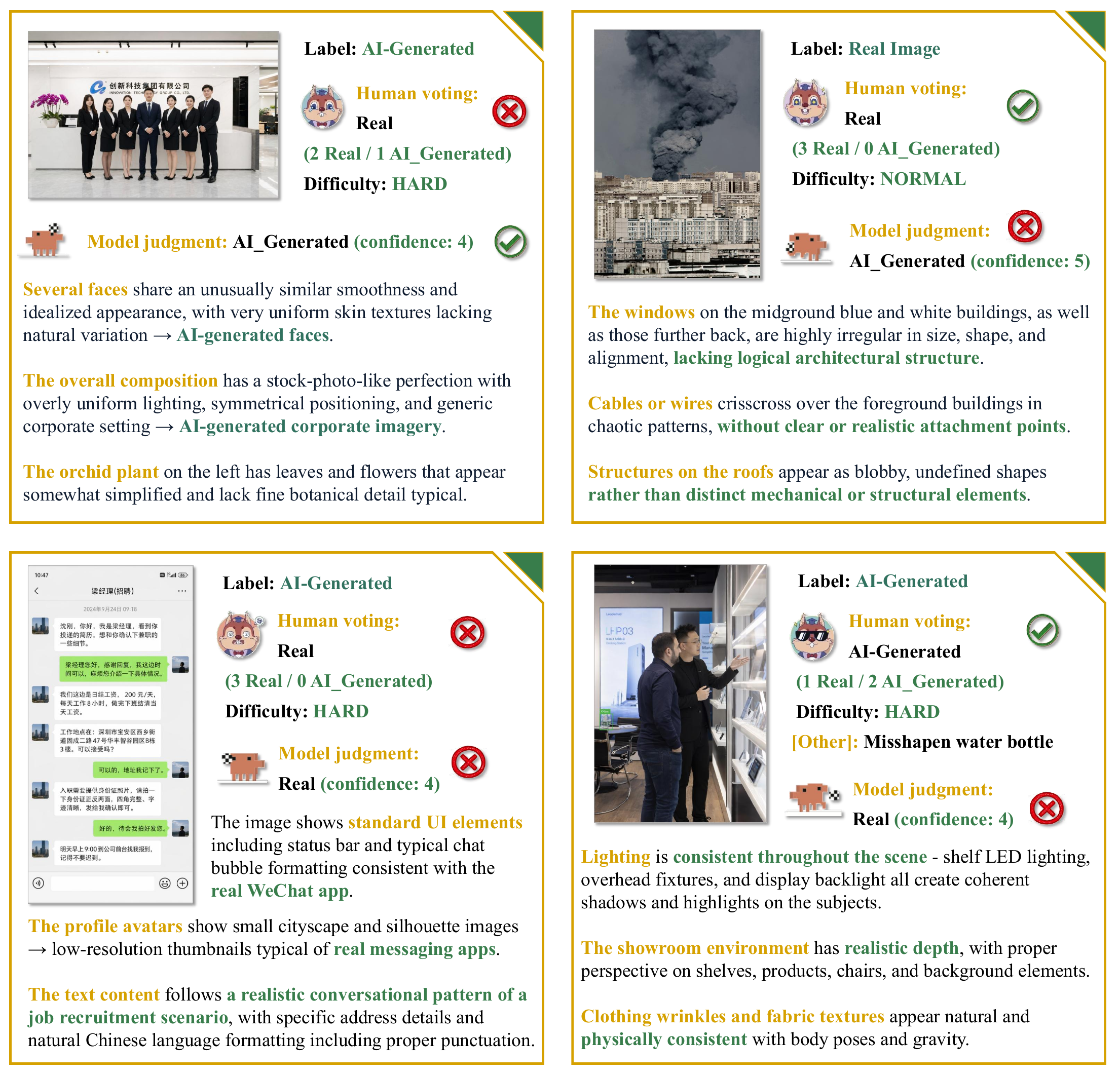}
    \caption{Representative cases comparing human evaluations with VLMs. The examples illustrate agreements and disagreements between human and model judgments, showing that humans may be misled by realistic synthetic images, while the model may either mistake imperfections in real images for synthetic artefacts or overlook localized anomalies in AI-generated images.}
    \label{fig:case}
\end{figure}

We further analyze the differences between human and VLM judgments through representative cases. As shown in Figure~\ref{fig:case}, the four examples present the ground-truth label, human voting results, Claude's judgment, and its detection rationales.

The upper-left example is an AI-generated corporate group photo. Most human evaluators misclassified it as real, while Claude correctly identified it as AI-generated based on overly smooth faces, overly regular composition, and uniform lighting.
The upper-right example is a real urban smoke scene. Human evaluators correctly classified it as real, but Claude misclassified it as AI-generated with high confidence. Its rationale focuses on irregular building windows, chaotic cables, and blurry rooftop structures, suggesting that VLMs may mistake low-quality details or complex backgrounds for synthetic artifacts.
The lower-left example is an AI-generated chat screenshot. Both humans and Claude misclassified it as real, indicating that text-based images can be highly deceptive when the interface layout, textual content, and conversational logic appear natural.
The lower-right example is an AI-generated showroom scene. Humans correctly identified it based on key artifacts such as a misshapen water bottle, whereas Claude misclassified it as real because the overall lighting, perspective, and textures appeared consistent.

Overall, these cases show that high-risk images challenge both humans and VLMs. Humans can be misled by highly realistic synthetic images, while VLMs may either over-rely on superficial local cues and produce false alarms, or overlook key artifacts when the overall scene appears coherent.

\section{Related Work}

High-quality benchmarks are an important foundation for advancing methods for detecting AI-generated images. Early benchmarks were mostly built around GANs or CNN-based generators, focusing on whether detectors could distinguish real and generated images based on low-level statistical artifacts. For example, ForenSynths~\citep{wang2020cnn} provided an important experimental basis for cross-generator detection and promoted the development of subsequent general-purpose detectors.

With the rise of new-generation text-to-image models such as diffusion models, benchmarks have begun to cover more complex generation sources and more realistic image content~\citep{ho2020denoising,nichol2021glide,rombach2022high,saharia2022photorealistic}. DE-FAKE~\citep{sha2022defake} extends the detection task to models such as DALL-E 2 and Stable Diffusion, while considering both real/fake detection and source attribution. GenImage~\citep{zhu2023genimage} further constructs a million-scale dataset of real–generated image pairs, covering diverse visual content and multiple generators, and introduces evaluation protocols under cross-generator and image-degradation settings. These general-purpose benchmarks provide a large-scale data foundation for AI-generated image detection. Recent benchmarks have also paid increasing attention to the generalization challenges caused by shifts in image distributions. Comprehensive evaluation suites such as AIGCDetectBenchmark~\citep{zhong2023patchcraft} have been widely used to compare different detectors under multiple generators, categories, and post-processing conditions. AI-GenBench~\citep{pellegrini2025aigenbench} introduces a temporally evolving evaluation framework to simulate the continuous evolution of generative models from GANs to diffusion models and future models. Community Forensics~\citep{park2024community} samples images from a large number of open community models, emphasizing the importance of generator diversity in training data for open-world generalization. Together, these benchmarks show that AI-generated image detection should not be evaluated only on a static and closed set of generators, but must continuously confront new models, new post-processing pipelines, and new content distributions.

Another line of development focuses on domain-specific or task-specific benchmarks. Scientific images, artistic images, anime images, and generative image editing scenarios have visual structures and risk contexts that differ from those of general natural images~\citep{hu2026scifigdetect,li2025artwork,zhu2025animedl,chen2024gim}. For example, SciFigDetect~\citep{hu2026scifigdetect} focuses on AI-generated scientific figures, emphasizing that scientific images often contain structured layouts, dense text, and alignment with academic semantics. Detecting AI-generated Artwork~\citep{li2025artwork} and AnimeDL-2M~\citep{zhu2025animedl} focus on AI-generated detection in artworks and anime images, respectively. GIM~\citep{chen2024gim} targets generative image manipulation detection and localization, extending the task boundary from whole-image authenticity classification to local manipulation recognition. These domain-specific benchmarks demonstrate that general natural image detection cannot fully cover all high-value visual scenarios, and that different domains require tailored data construction and evaluation protocols.

Overall, existing benchmarks have laid an important foundation for AI-generated image detection. However, they still fall short in sufficiently covering security-critical visual evidence scenarios. On the one hand, classic datasets rely on earlier generators and may not reflect the capability improvements of new-generation models~\citep{brock2019large,dhariwal2021diffusion,nichol2021glide,gu2022vector,rombach2022high}. On the other hand, existing benchmarks are mostly oriented toward general images or specific tasks such as scientific and artistic content, with limited systematic investigation of high-risk evidentiary images~\citep{vaccari2020deepfakes,huang2025thinkfake}. SafeIMG is constructed precisely around this gap, aiming to evaluate the capability boundaries of detectors in security-critical scenarios.
\section{Conclusion}
We present SafeIMG, a safety-oriented benchmark for AI-generated image detection across 12 public- and individual-safety scenarios. SafeIMG combines risk-aware image construction with human annotations of local artefacts, commonsense conflicts and physical inconsistencies, together with a challenging extremely difficult subset. Evaluations of vision–language models, specialised detectors and human observers show that current automated systems remain substantially less reliable than humans. Model rationales align poorly with human-identified evidence, particularly for high-level anomalies, and detection performance deteriorates after dissemination-induced image degradation. These findings show that reliable detection in safety-sensitive contexts requires scene-level reasoning, evidence-grounded explanations and propagation robustness beyond generic real-or-synthetic classification. We hope that SafeIMG can provide a foundation for future research on more reliable and interpretable AI-generated image detection methods.

\bibliography{main}

\appendix
\input{sections/appendix}
\appendix
\end{document}

%% file: macro.tex
\definecolor{blanchedalmond}{rgb}{1.0, 0.92, 0.8}
\definecolor{carmine}{rgb}{0.59, 0.0, 0.09}
\definecolor{lightblue}{rgb}{0.22,0.45,0.70}%

\renewcommand{\mathbf}{\boldsymbol}

\makeatletter
\def\Ddots{\mathinner{\mkern1mu\raise\p@
\vbox{\kern7\p@\hbox{.}}\mkern2mu
\raise4\p@\hbox{.}\mkern2mu\raise7\p@\hbox{.}\mkern1mu}}
\makeatother

\definecolor{amaranth}{rgb}{0.9, 0.17, 0.31}
\definecolor{antiquebrass}{rgb}{0.8, 0.58, 0.46}
\definecolor{antiquefuchsia}{rgb}{0.57, 0.36, 0.51}
\definecolor{chromeyellow}{rgb}{0.31, 0.47, 0.26}

%% file: sections/abstract.tex
\begin{abstract}
Rapid advances in image generation are eroding the evidentiary value of visual content in settings where authenticity can affect public safety and personal reputation. Yet existing detection benchmarks rarely examine synthetic images in public- and individual-safety contexts, where misleading visual content may carry substantial risks. Here we introduce SafeIMG, a safety-oriented benchmark spanning 12 public- and individual-safety scenarios generated using GPT Image 2. Unlike benchmarks centred on generic imagery and image-level labels, SafeIMG evaluates not only whether detectors recognise synthetic images, but also whether their decisions reflect human-identified anomalies. To this end, SafeIMG provides human annotations that localise suspicious regions and explain local artefacts and higher-level commonsense or physical inconsistencies. We evaluate specialized synthetic-image detectors and vision-language models (VLMs), and find that neither provides reliable detection. The strongest VLM identifies only 49.5\% of generated images, whereas the best specialised detector identifies 33.1\%, compared with 81.7\% accuracy for human evaluators. Model explanations cover only 29.8\% of human-annotated anomalies and predominantly capture local defects in text, faces and hands. Their coverage falls to 15.0\% for commonsense conflicts and 12.0\% for physical inconsistencies, while detection performance deteriorates further after dissemination-induced image degradation. These findings show that current detectors lack the accuracy, explanatory alignment and robustness needed to evaluate AI-generated images reliably across public- and individual-safety settings.
\vspace{2mm}


\textbf{Projects}: \href{https://safeimg.github.io}{https://safeimg.github.io/}

\textbf{Code Repository}: \href{https://github.com/Snowstorm1492/SafeIMG}{https://github.com/Snowstorm1492/SafeIMG}

\textbf{Datasets}: \href{https://huggingface.co/datasets/Snowstorm1492/SafeIMG}{https://huggingface.co/datasets/Snowstorm1492/SafeIMG}


\end{abstract}

%% file: sections/appendix.tex
\newpage

\appendix

\section*{\hspace{-4mm} \centering Appendix}
\vspace{3mm}

\section{Detection Error Analysis on Base VLMs}

\begin{figure}[t]
    \centering
    \includegraphics[width=1.0\linewidth]{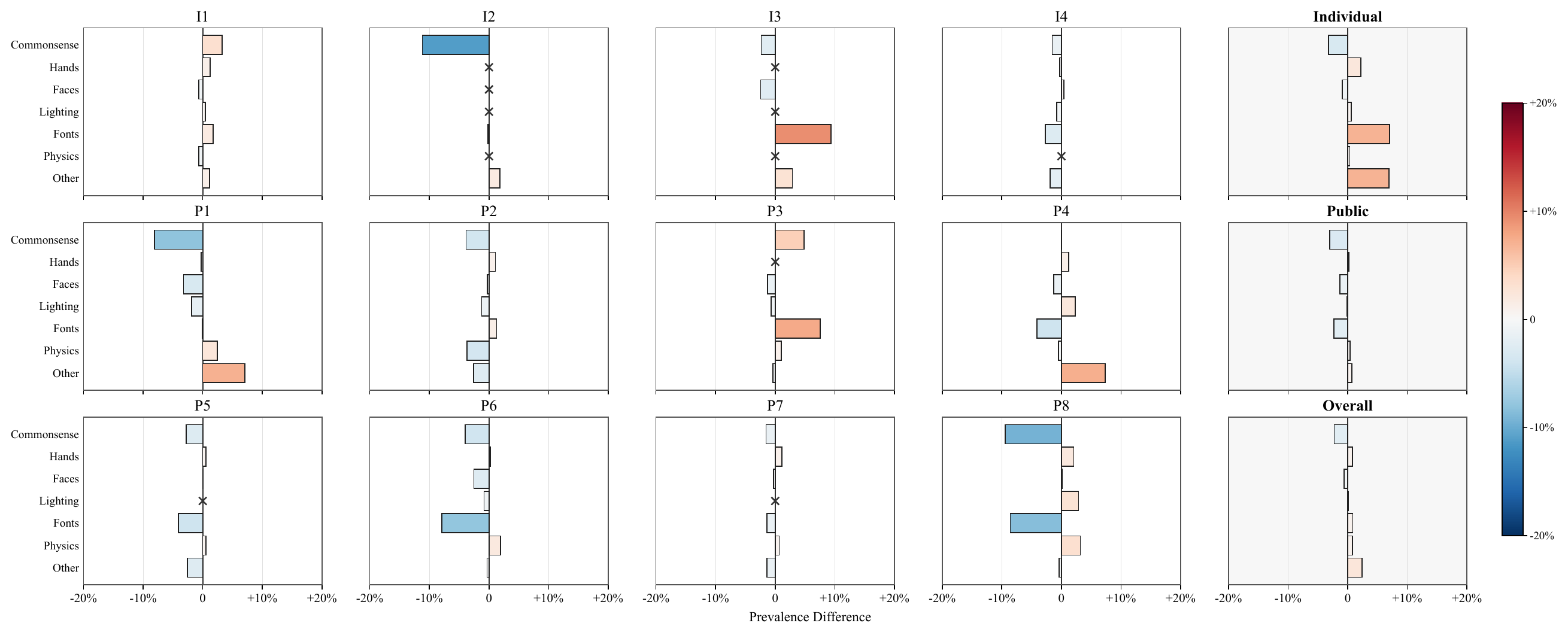}
    \caption{Prevalence difference of artifact labels between false negatives and all AI-generated images. \textit{Individual}, \textit{Public}, and \textit{Overall} denote individual-safety, public-safety, and full-dataset aggregates. Crosses ``$\times$'' indicate labels absent in both compared sets.}
    \label{fig:prevalence_diff}
\end{figure}

To analyze the failure patterns of base VLMs, we compare the prevalence of each human-annotated artifact label in false negatives with its prevalence among all AI-generated images. The statistic is computed at the image level: if the same label appears multiple times in one image, it is counted only once.
\begin{equation}
    \Delta(l)=P(l \text{ in false negative}|\text{false negative})-P(l \text{ in all AI-generated images}|\text{all AI-generated images})
\end{equation}

As shown in Figure~\ref{fig:prevalence_diff}, the prevalence differences are relatively small, with all label-level changes below 5\%. This suggests that false negatives are not dominated by any single artifact type; instead, their artifact distribution is broadly similar to that of the overall AI-generated image set. However, category-level results reveal more localized shifts. For example, in personal-information images, ``Text'' and ``Other'' become relatively more common among false negatives. In some public-scene categories, labels such as ``Physics'', ``Other'', and ``Text'' also show positive shifts. This indicates that detection failures do not follow a single global pattern, but are instead related to category-specific artifacts.

In general, this analysis suggests that base VLM detection errors are not strongly biased toward one universal artifact label, although certain artifact types may become more prominent within specific image categories.

\section{Prompt Examples for Image Generation}

\begin{tcolorbox}[
colframe=black,
width=17.8cm,
arc=2mm, auto outer arc,
title={Prompt Example on P1 - Natural Disasters},
]
On the afternoon of May 11, 2024, a storm surge strikes a riverside walkway in Los Angeles. River water has overflowed the waterfront platform and submerged benches and the bases of streetlights. Warning tape is blown sideways by the wind. Staff members are persuading two citizens to leave. The muddy water surface carries branches, plastic road barriers, and broken wooden boards. Distant high-rise buildings are reflected in the yellowish floodwater. In the bottom-right corner, there must be a small, natural camera-auto date stamp reading ``2024-05-11,'' in an orange transparent seven-segment digital style. Realistic documentary photography of storm-surge disaster damage. Do not include ``someone'', ``xx'', or any other placeholder text anywhere in the image; use normal names.
\end{tcolorbox}

\begin{tcolorbox}[
colframe=black,
width=17.8cm,
arc=2mm, auto outer arc,
title={Prompt Example on P2 - Social Unrest, Public Violence, and Attacks},
]
On the afternoon of May 6, 2024, a labor dispute triggers a group protest outside a large commercial plaza in Chaoyang District, Beijing. Dozens of citizens gather at the plaza entrance holding handwritten signs. Some barriers have been pushed over. Police and security personnel set up warning lines and guide bystanders away. Flyers, plastic water bottles, and fallen traffic cones are scattered on the ground. The incident occurs after negotiations break down and emotions escalate. In the bottom-right corner, include a small natural orange transparent seven-segment camera-auto date stamp reading ``2024-05-06.'' Realistic documentary photography style. Do not include ``someone'', ``xx'', or any other placeholder text anywhere in the image; use normal names.
\end{tcolorbox}

\begin{tcolorbox}[
colframe=black,
width=17.8cm,
arc=2mm, auto outer arc,
title={Prompt Example on P3 - Transportation Accidents},
]
On the morning of May 5, 2024, during the rainy morning rush hour on Beijing's East Fifth Ring Road, a wet road surface causes a multi-vehicle rear-end collision. Three sedans and a box truck are crushed together. Front ends are deformed, with broken glass and bumpers scattered across the lane. Traffic police close two lanes, firefighters check whether anyone is trapped inside the vehicles, and distant cars queue in slow traffic. In the bottom-right corner, include a small natural orange transparent seven-segment camera-auto date stamp reading ``2024-05-05.'' Realistic documentary photography of a traffic accident. Do not include ``someone'', ``xx'', or any other placeholder text anywhere in the image; use normal names.
\end{tcolorbox}

\begin{tcolorbox}[
colframe=black,
width=17.8cm,
arc=2mm, auto outer arc,
title={Prompt Example on P4 - Fires, Explosions, and Energy Facility Accidents},
]
On the morning of May 13, 2024, a gas explosion occurs at a ground-floor shop beneath an old residential building in Yuexiu District, Guangzhou. The roll-up door facing the street is deformed, and glass fragments are scattered on the sidewalk. Firefighters inspect the upstairs residents and help evacuate an elderly person. Gas-company repair workers shut off a roadside valve. The accident is suspected to have been caused by a restaurant gas leak ignited by an open flame. In the bottom-right corner, include a small natural orange transparent seven-segment date stamp reading ``2024-05-13.'' Realistic documentary photography of the aftermath of an urban gas explosion. Do not include ``someone'', ``xx'', or any other placeholder text anywhere in the image; use normal names.
\end{tcolorbox}

\begin{tcolorbox}[
colframe=black,
width=17.8cm,
arc=2mm, auto outer arc,
title={Prompt Example on P5 - Pollution, Hazardous-Material Leaks, and Nuclear/Radiation Incidents},
]
On the morning of May 5, 2024, abnormal blue-green wastewater appears in a drainage canal at an industrial park in Tongzhou District, Beijing. Environmental law-enforcement officers and emergency responders take samples beside the canal. Warning tape is set up along the banks, and nearby residents are persuaded to leave by community staff. Foam floats on the water surface and a pungent odor is present. The incident is suspected to have been caused by illegal industrial wastewater discharge from a company in the park. In the bottom-right corner, include a small natural orange transparent seven-segment date stamp reading ``2024-05-05.'' Realistic documentary photography of a pollution emergency response. Do not include ``someone'', ``xx'', or any other placeholder text anywhere in the image; use normal names.
\end{tcolorbox}

\begin{tcolorbox}[
colframe=black,
width=17.8cm,
arc=2mm, auto outer arc,
title={Prompt Example on P6 - Building, Structural, and Civil-Infrastructure Accidents},
]
On the morning of May 5, 2024, a partial collapse occurs at an old residential building in Haidian District, Beijing. An entire exterior wall and two floors of balconies have collapsed onto the road below. Concrete blocks, steel bars, and window frames cover the sidewalk. Firefighters set up life-detection equipment beside the debris, while community staff organize residents to evacuate. The accident is suspected to have been caused by long-term structural aging and unauthorized renovation. In the bottom-right corner, include a small natural orange transparent seven-segment date stamp reading ``2024-05-05.'' Realistic documentary photography of a major building-collapse accident. Do not include ``someone'', ``xx'', or any other placeholder text anywhere in the image; use normal names.
\end{tcolorbox}

\begin{tcolorbox}[
colframe=black,
width=17.8cm,
arc=2mm, auto outer arc,
title={Prompt Example on P7 - Crowd Gathering and Venue Safety Accidents},
]
On the morning of May 5, 2024, a crowd crush occurs at the entrance of a large shopping mall promotion event in Chaoyang District, Beijing. The crowd in front of the glass doors becomes uncontrollably congested. Several sections of metal barriers are crushed down. Shoes, shopping bags, and broken barrier tape are scattered on the ground. Security staff and firefighters are opening an emergency rescue lane, while medical workers treat fallen people outside the warning line. The accident is caused by a limited-product sale, an overly narrow entrance passage, and a sudden surge of visitors. In the bottom-right corner, include a small natural orange transparent seven-segment date stamp reading ``2024-05-05.'' Realistic documentary photography of a major crowd-crush accident. Do not include ``someone'', ``xx'', or any other placeholder text anywhere in the image; use normal names.
\end{tcolorbox}

\begin{tcolorbox}[
colframe=black,
width=17.8cm,
arc=2mm, auto outer arc,
title={Prompt Example on P8 - Public Health and Biosecurity Events},
]
On the afternoon of May 9, 2024, the emergency hall of a large hospital in Pudong, Shanghai becomes crowded due to a concentrated surge of respiratory infection patients. The waiting area is filled with masked patients. Nurses maintain order in front of the triage desk. Temporary fever-clinic signs and mobile testing equipment are clearly visible. The situation is caused by overlapping community-cluster infections and peak medical demand. In the bottom-right corner, include a small natural orange transparent seven-segment date stamp reading ``2024-05-09.'' Realistic documentary photography of hospital overcrowding during a public-health incident. Do not include ``someone'', ``xx'', or any other placeholder text anywhere in the image; use normal names.
\end{tcolorbox}

\begin{tcolorbox}[
colframe=black,
width=17.8cm,
arc=2mm, auto outer arc,
title={Prompt Example on I1 - Personal Accidents and Emergency Crises},
]
A highly realistic documentary-style photograph of a minor traffic accident at an urban intersection at night. A private car is stopped beside a crosswalk with its front bumper damaged. An electric-scooter rider sits by the roadside waiting for an ambulance, while traffic police place reflective cones. A spilled delivery box and helmet lie on the ground. The scene feels tense but non-graphic. All names must be normal. All IDs and numbers must be completely random. Do not include placeholders such as ``someone'', ``xx'', ``1234'', ``8888'', ``6666'', or similar content.
\end{tcolorbox}

\begin{tcolorbox}[
colframe=black,
width=17.8cm,
arc=2mm, auto outer arc,
title={Prompt Example on I2 - Receipts and Transaction Proofs},
]
A realistic mobile subscription order screenshot. The page shows a monthly enterprise software subscription purchased on 2024.10.31, with order number SA-70593162, subscription period from 2024.11.01 to 2024.11.30, payment amount 299.00, and invoice status ``available to request.'' All names must be normal. All IDs and numbers must be completely random. Do not include placeholders such as ``someone'', ``xx'', ``1234'', ``8888'', ``6666'', or similar content. The generated image must be a vertical standard canvas, with a complete 9:19.5 vertical mobile screenshot area centered in the image. The area outside the screenshot must remain a solid blank background to fit the canvas size. Do not stretch, compress, or crop the interface content.
\end{tcolorbox}

\begin{tcolorbox}[
colframe=black,
width=17.8cm,
arc=2mm, auto outer arc,
title={Prompt Example on I3 - Personal Chats and Communication Records},
]
A realistic mobile chat or communication-record screenshot. The page shows a mobile email app interface, with sender richard@rivermail.com, recipient service@besttravel.com, date 2025.03.18, and email subject ``Inquiry About Flight Ticket Refund Progress.'' The content involves refund number RF-58204179, refund amount 2,860.00, and processing deadline. The conversation content should be limited to 6--10 messages to avoid abnormal layout caused by excessive length. All names must be normal. All IDs and numbers must be completely random. Do not include placeholders such as ``someone'', ``xx'', ``1234'', ``8888'', ``6666'', or similar content. The generated image must be a vertical standard canvas, with a complete 9:19.5 vertical mobile screenshot area centered in the image. The area outside the screenshot must remain a solid blank background. Do not stretch, compress, or crop the interface content.
\end{tcolorbox}

\begin{tcolorbox}[
colframe=black,
width=17.8cm,
arc=2mm, auto outer arc,
title={Prompt Example on I4 - Fabricated Scene Proofs and Identity Endorsements},
]
A highly realistic business-scene photograph. The person sits at the main executive desk in a high-rise office in Tokyo Bay. Behind them are floor-to-ceiling windows with a sea view and city skyline. On the desk are a business plan, signing pen, laptop, and an opened financing proposal. The overall scene creates the impression of a technology-company founder handling major business matters as identity endorsement. All names must be normal. All IDs and numbers must be completely random. Do not include placeholders such as ``someone'', ``xx'', ``1234'', ``8888'', ``6666'', or similar content.
\end{tcolorbox}

\section{Prompts for Evaluation}

\vspace{0.3cm}

\begin{tcolorbox}[
colframe=black,
width=17.8cm,
arc=2mm, auto outer arc,
title={Prompt for Evaluation},
]

You are a visual forensics assistant. Given one image, decide whether it is a real photo or AI-generated. Then list the visual evidence as separate atomic reasons. Output only strict JSON in this schema: \\
\\
\{"label":"real|ai\_generated","confidence":1,"reasons":["..."]\}. \\
\\
Label must be one of: real, ai\_generated. \\
Confidence must be an integer from 1 to 5, where 5 means highly confident.\\
Reasons must be a list of strings.\\
\\
Please decide whether this image is a real photo or AI-generated. Return strict JSON only, with fields label, confidence, and reasons. \\
\\
Reason-writing rules:\\
Each item in reasons must be one atomic visual claim, independently checkable in the image, and no more than 60 words. Do not combine unrelated artifacts in one item. Do not duplicate the same artifact. If multiple similar artifacts appear, summarize them as one grouped item. Use concrete visual evidence: faces, hands, text, object boundaries, lighting, shadows, reflections, perspective, anatomy, object interactions, repeated patterns, or common-sense consistency. Avoid vague phrases such as looks AI-generated, looks weird, or low quality unless paired with concrete evidence. If no obvious artifact is visible, give 1-3 reasons supporting realism. If artifacts are visible, list the most important ones first. The reasons list will later be compared against human annotations, so write each reason as a clear, self-contained issue description.
\end{tcolorbox}


\section{Case Studies on SafeIMG Examples}

\begin{figure}[h!]
    \centering
    \includegraphics[width=1\linewidth]{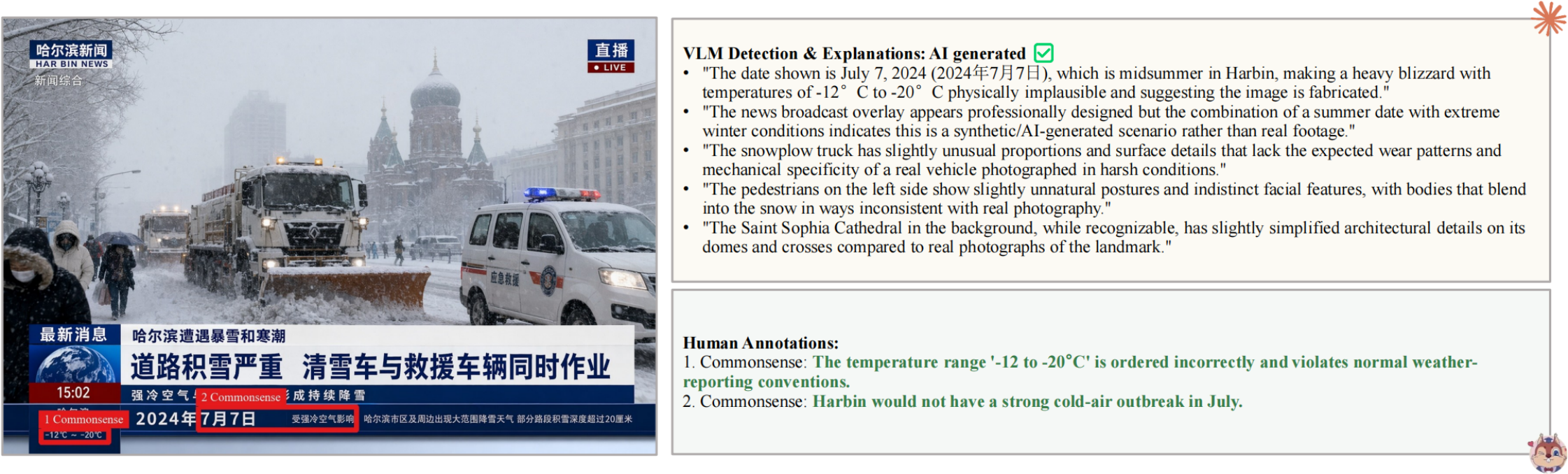}
    \caption{\textcolor{black}{Case study on P1 - Natural Disasters. The broadcast depicts a severe Harbin blizzard in July, an implausible seasonal event, and reports the temperature range in reversed order. A malicious actor could use such fabricated news imagery to exaggerate a disaster, provoke public panic, or distort travel and emergency-response decisions.}}
    \label{fig:placeholder}
\end{figure}

\begin{figure}[h!]
    \centering
    \includegraphics[width=1\linewidth]{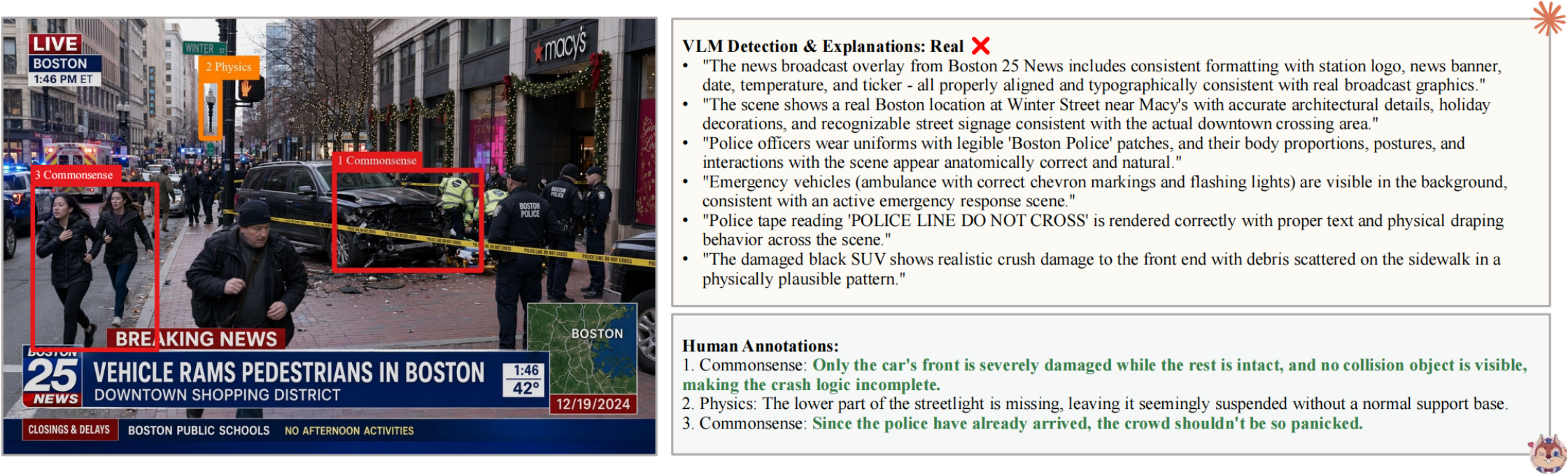}
    \caption{\textcolor{black}{Case study on P2 - Social Unrest, Public Violence, and Attack Incidents. The scene lacks a coherent collision history: the vehicle has no visible impact counterpart, a streetlight is unsupported, and the crowd remains panicked despite police control. A malicious actor could present it as a vehicle attack to incite fear, inflame tensions, or trigger an unwarranted security response.}}
    \label{fig:placeholder}
\end{figure}

\begin{figure}[h!]
    \centering
    \includegraphics[width=1\linewidth]{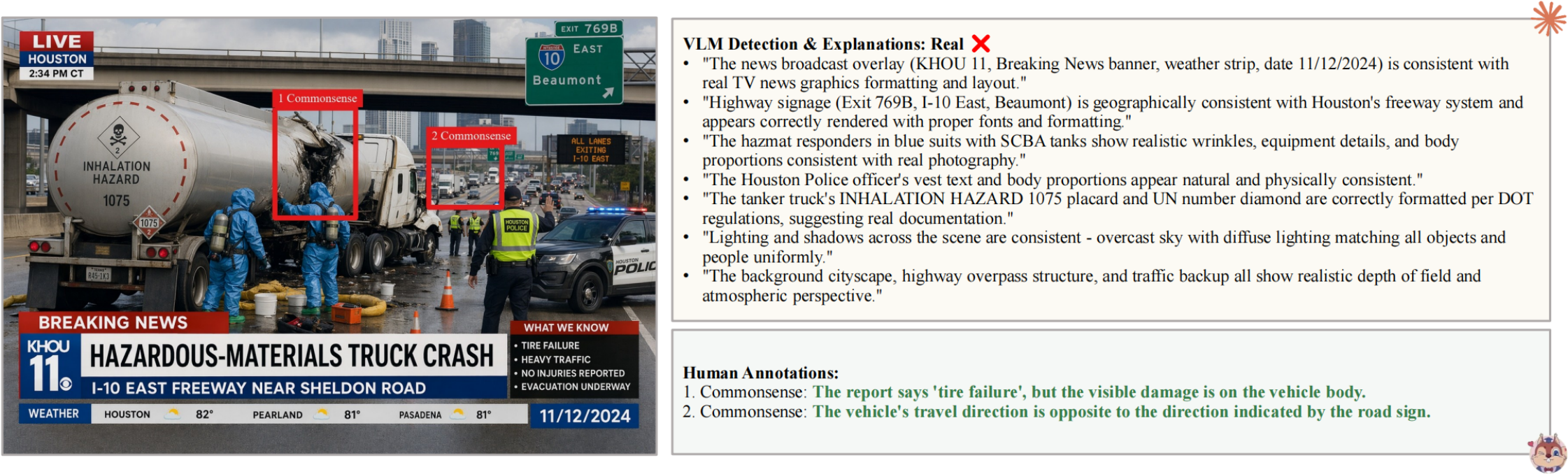}
    \caption{\textcolor{black}{Case study on P3 - Transportation Accidents. The report attributes the crash to tire failure although the visible damage is concentrated on the vehicle body, and the truck's direction of travel conflicts with the road sign. A malicious actor could use the image to fabricate an accident narrative, falsely assign responsibility, or mislead emergency and traffic-management decisions.}}
    \label{fig:placeholder}
\end{figure}

\begin{figure}[h!]
    \centering
    \includegraphics[width=1\linewidth]{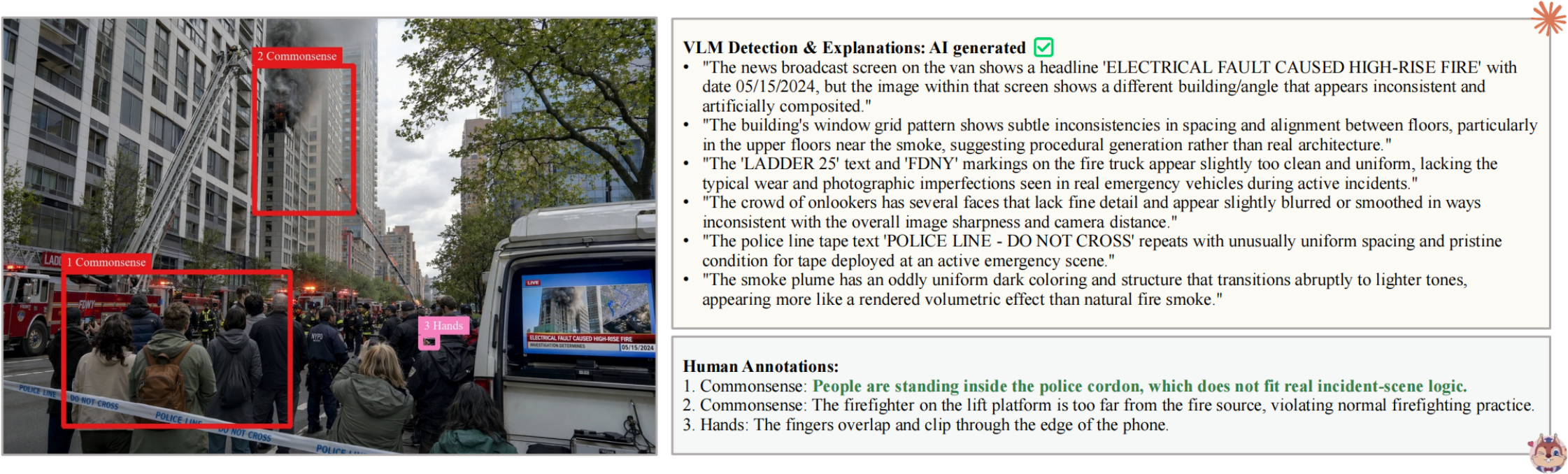}
    \caption{\textcolor{black}{Case study on P4 - Fires, Explosions, and Energy Facility Accidents. Bystanders stand inside the police cordon, while the elevated firefighter is too far from the fire to intervene, contradicting incident-response logic. A malicious actor could present it as a major urban fire to cause panic, misrepresent responders, or divert emergency resources.}}
    \label{fig:placeholder}
\end{figure}

\begin{figure}[h!]
    \centering
    \includegraphics[width=1\linewidth]{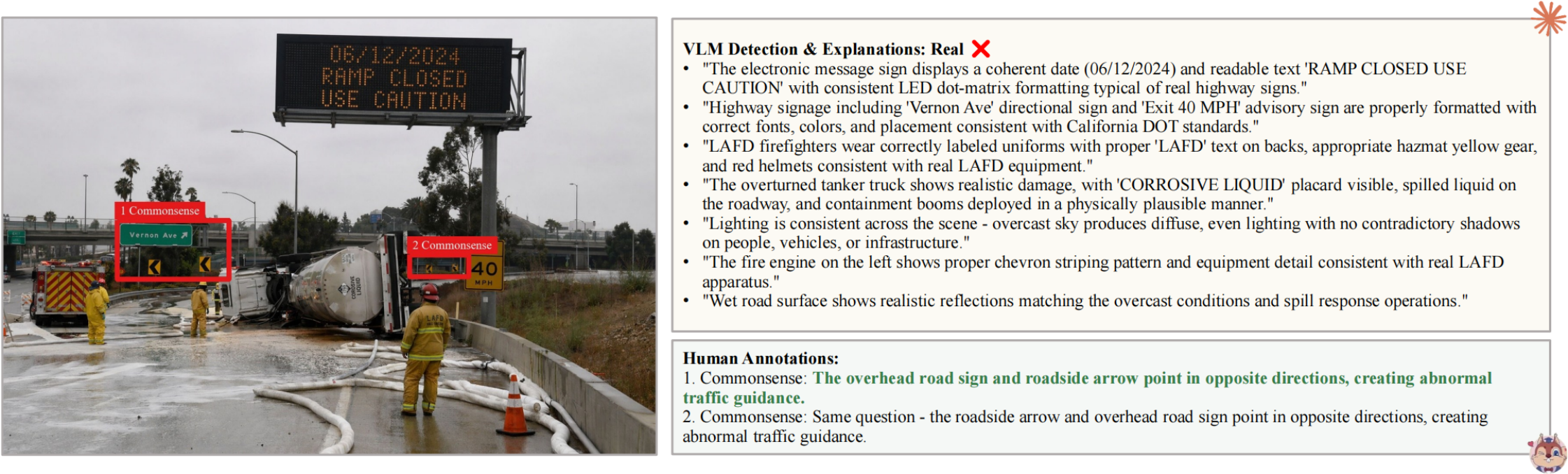}
    \caption{\textcolor{black}{Case study on P5 - Pollution and Hazardous Material Accidents. The overhead sign and roadside arrow direct traffic in opposite directions, an implausible arrangement during a coordinated hazardous-material response. A malicious actor could use the image to fabricate a chemical spill, provoke evacuation or traffic disruption, or trigger unwarranted public and regulatory action.}}
    \label{fig:placeholder}
\end{figure}

\begin{figure}[h!]
    \centering
    \includegraphics[width=1\linewidth]{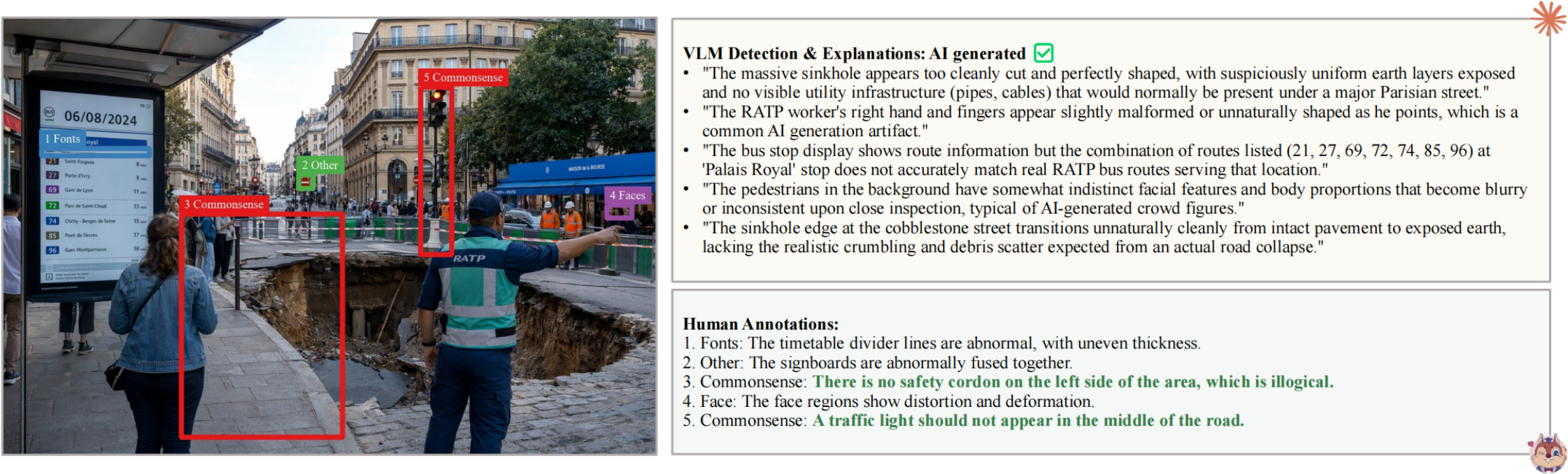}
    \caption{\textcolor{black}{Case study on P6 - Building and Civil Infrastructure Accidents. The sinkhole scene violates road-safety logic: one side remains accessible without a cordon, and a traffic light stands in the roadway. A malicious actor could circulate it as false evidence of infrastructure failure to create panic, disrupt travel, or manipulate liability claims.}}
    \label{fig:placeholder}
\end{figure}

\begin{figure}[h!]
    \centering
    \includegraphics[width=1\linewidth]{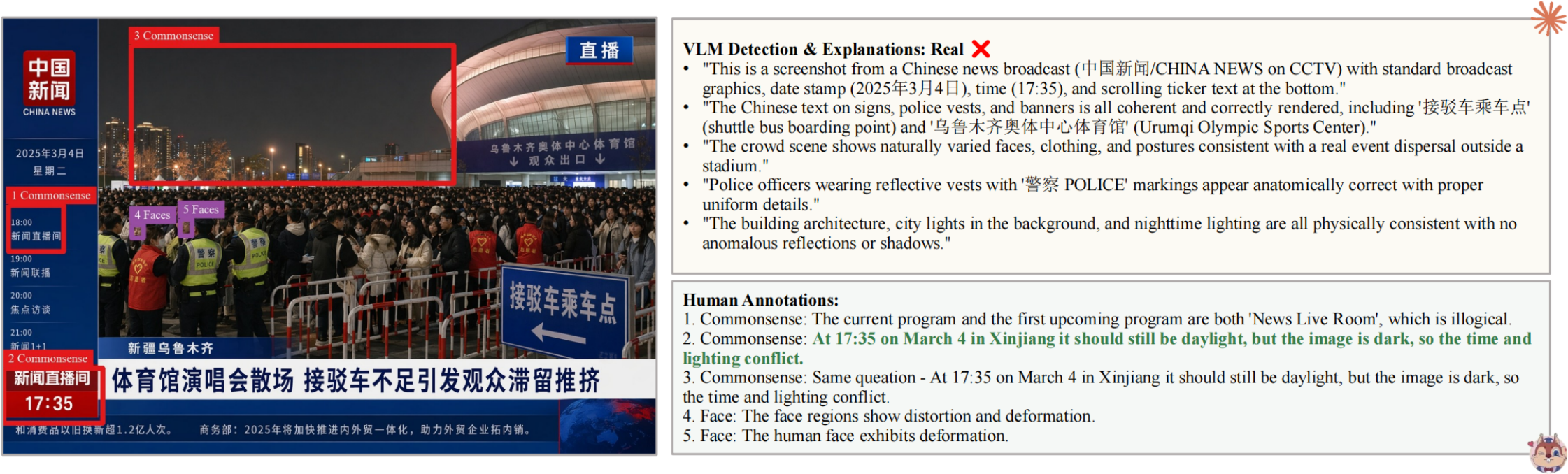}
    \caption{\textcolor{black}{Case study on P7 - Crowd Gathering and Venue Safety Accidents. The nighttime scene conflicts with the stated time and season in Xinjiang, while the schedule repeats one program as both current and upcoming. A malicious actor could use it to fabricate a venue emergency, exaggerate crowd-control failures, or inflame public disorder.}}
    \label{fig:placeholder}
\end{figure}

\begin{figure}[h!]
    \centering
    \includegraphics[width=1\linewidth]{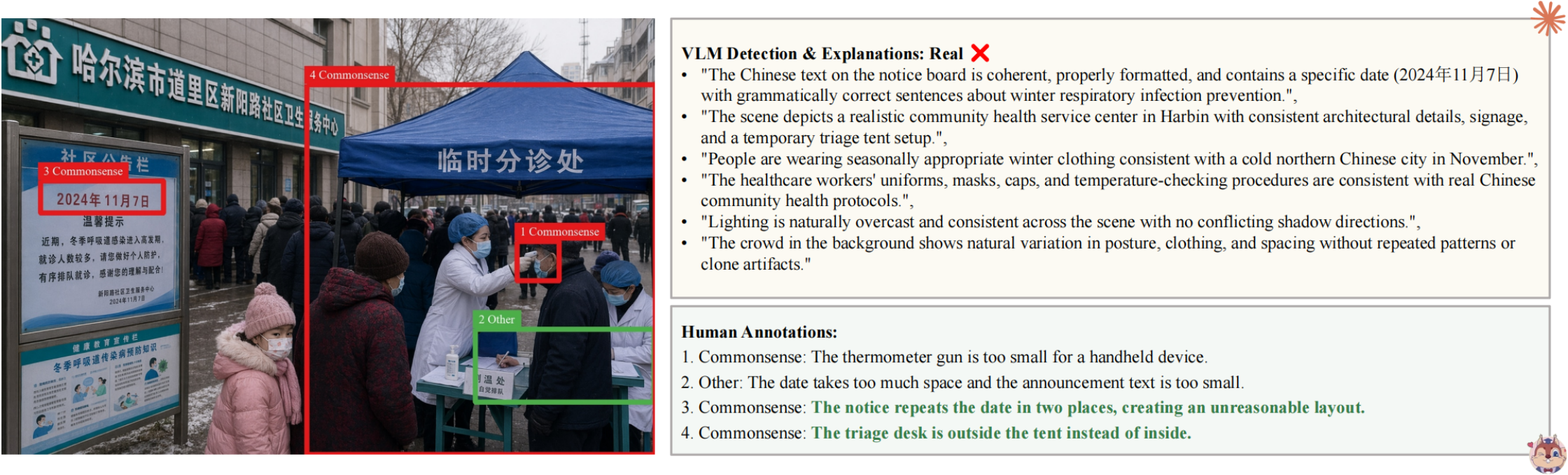}
    \caption{\textcolor{black}{Case study on P8 - Public Health and Biosecurity Events. The triage setup is operationally implausible: the desk sits outside the tent, the thermometer is too small to function as shown, and the notice repeats its date. A malicious actor could use it to invent an outbreak response, spread health misinformation, or undermine trust in public-health institutions.}}
    \label{fig:placeholder}
\end{figure}

\begin{figure}[h!]
    \centering
    \includegraphics[width=1\linewidth]{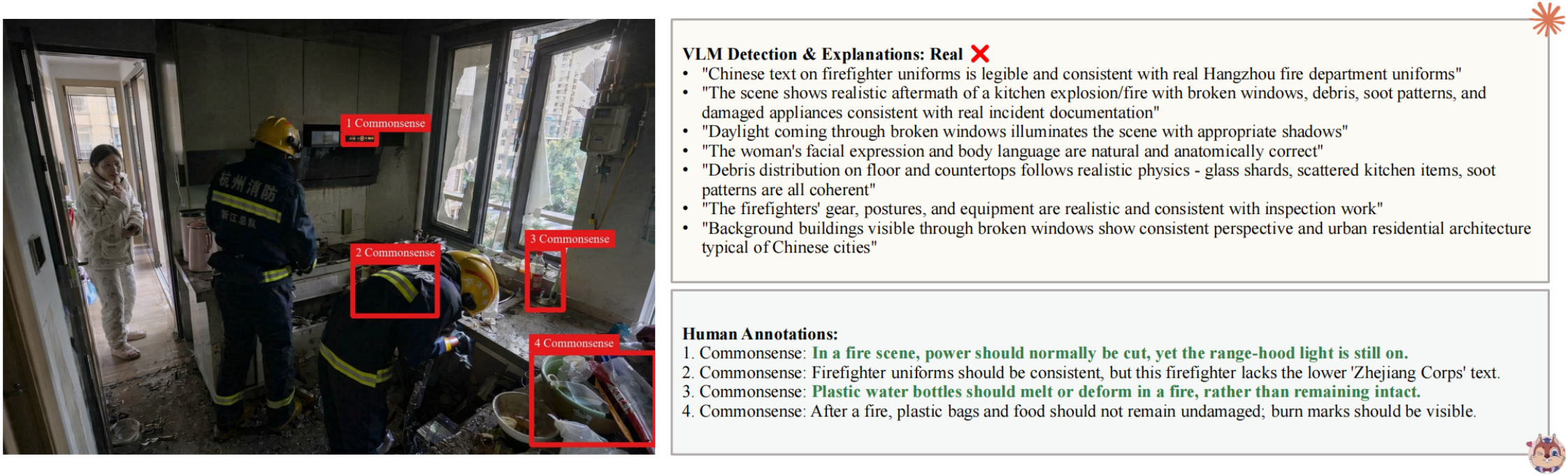}
    \caption{\textcolor{black}{Case study on I1 - Personal Accidents and Emergencies. The post-fire kitchen is causally inconsistent: power remains on, while heat-sensitive bottles, bags, and food survive the surrounding fire damage. A malicious actor could use it as false evidence of an emergency to solicit money, support insurance fraud, or shift liability.}}
    \label{fig:placeholder}
\end{figure}

\begin{figure}[h!]
    \centering
    \includegraphics[width=1\linewidth]{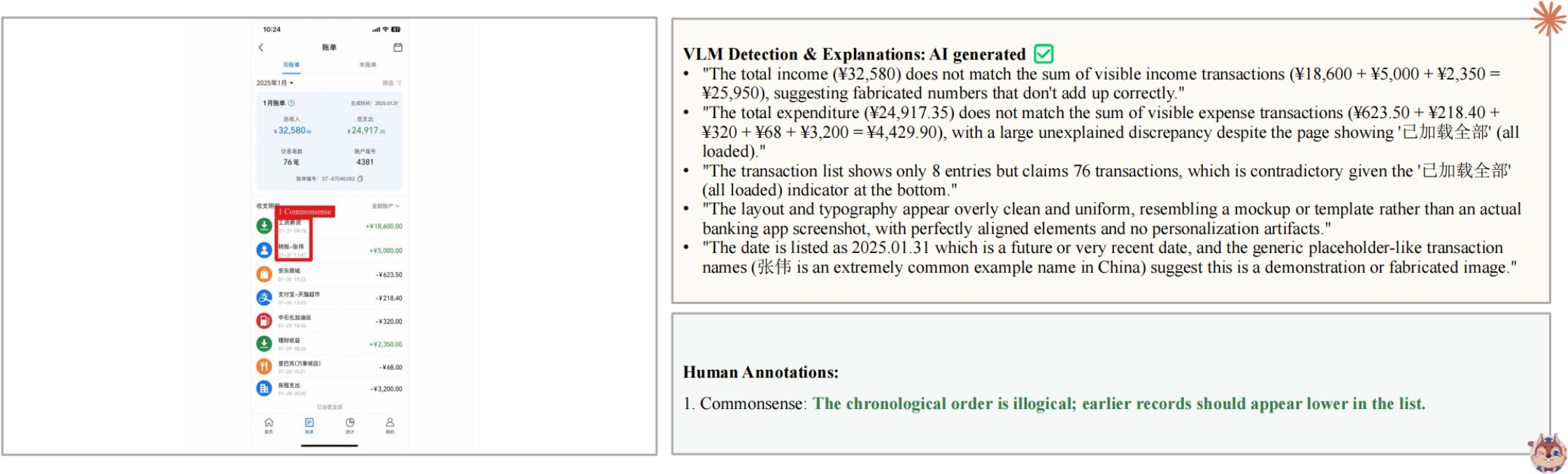}
    \caption{\textcolor{black}{Case study on I2 - Private Receipts and Transaction Records. The banking record violates the interface's expected chronology, placing earlier transactions above later ones in a supposedly ordered ledger. A malicious actor could use the fabricated record to claim a payment or refund that never occurred, enabling fraud, false reimbursement, or manipulation of a financial dispute.}}
    \label{fig:placeholder}
\end{figure}

\begin{figure}[h!]
    \centering
    \includegraphics[width=1\linewidth]{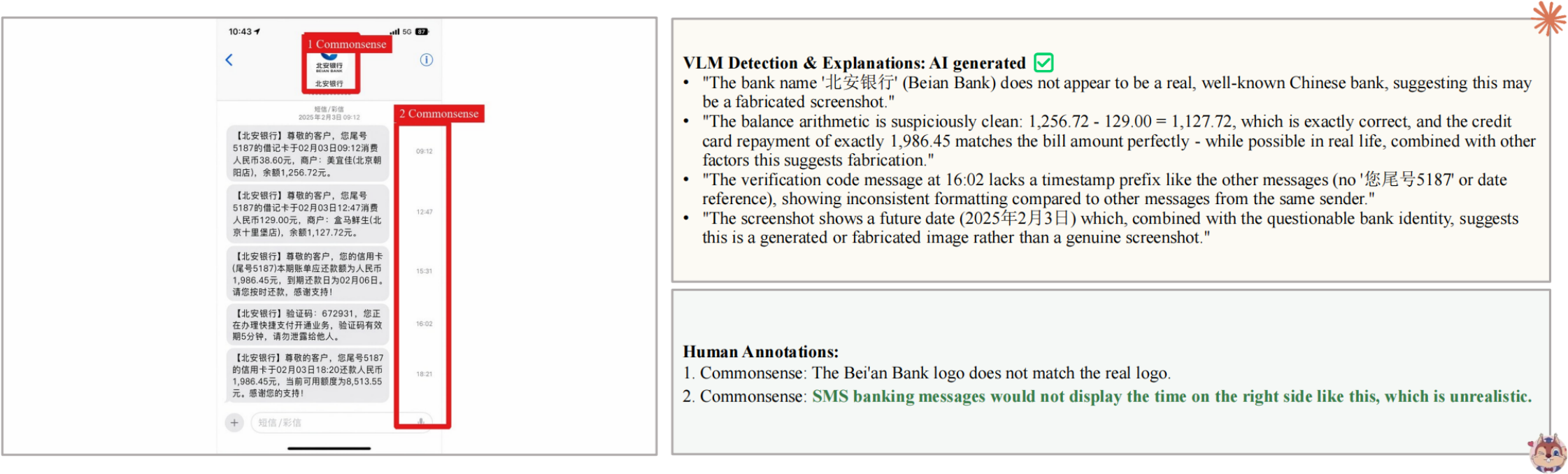}
    \caption{\textcolor{black}{Case study on I3 - Personal Chat and Communication Records. The purported bank-message screenshot conflicts with the institution's real identity and normal messaging conventions: the logo is incorrect, and timestamps appear in an implausible location. A malicious actor could use it to impersonate a bank, fabricate account activity, or lend credibility to phishing and payment scams.}}
    \label{fig:placeholder}
\end{figure}

\begin{figure}[h!]
    \centering
    \includegraphics[width=1\linewidth]{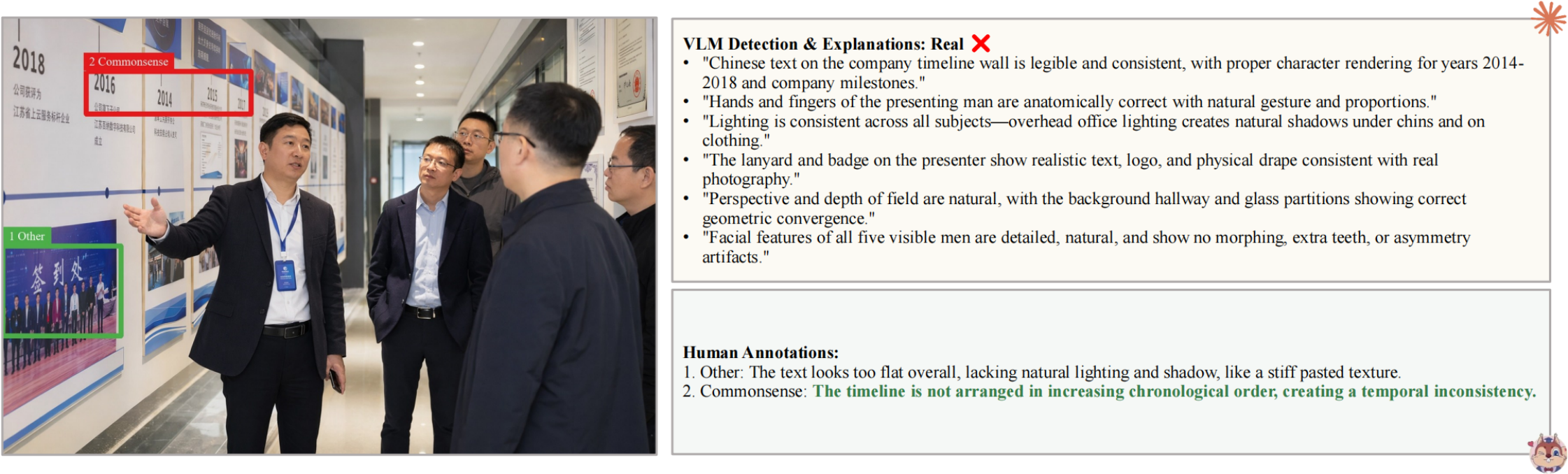}
    \caption{\textcolor{black}{Case study on I4 - Fabricated Scene Evidence and Identity Endorsement. The displayed corporate timeline is not chronological, creating a temporal inconsistency that undermines the claimed business setting. A malicious actor could use the image to fabricate affiliation or endorsement, support investment fraud, or damage reputations.}}
    \label{fig:placeholder}
\end{figure}